\RequirePackage{fix-cm}
\documentclass[twocolumn]{svjour3}          
\smartqed  

\usepackage{times}
\usepackage{epsfig}
\usepackage{graphicx}
\usepackage{graphics}
\usepackage{amsmath}
\usepackage{amssymb}
\usepackage{caption} 
\usepackage[export]{adjustbox}
\usepackage{amsmath}
\usepackage{comment}
\usepackage[colorinlistoftodos]{todonotes}
\usepackage{cite}

\usepackage{tabularx}
\usepackage{verbatim}

\usepackage{subfigure}
\usepackage{booktabs}
\usepackage{rotating}
\usepackage{multirow}
\usepackage{color}
\usepackage{xspace}

\usepackage{enumitem}

\usepackage{hyperref}
\hypersetup{
    colorlinks=true,
    linkcolor=blue,
    filecolor=magenta,      
    urlcolor=red,
}
\makeatletter
\newcommand\footnoteref[1]{\protected@xdef\@thefnmark{\ref{#1}}\@footnotemark}
\makeatother

\makeatletter
\DeclareRobustCommand\onedot{\futurelet\@let@token\@onedot}
\def\@onedot{\ifx\@let@token.\else.\null\fi\xspace}

\makeatother

\journalname{International Journal of Computer Vision - Special Issue on Deep Learning for Face Analysis}
\begin{document}

\title{Deep Affect Prediction in-the-wild: Aff-Wild Database and Challenge, Deep Architectures, and Beyond
}

\author{Dimitrios Kollias $^\star$        \and
        Panagiotis Tzirakis $^\dagger$\and
        Mihalis A. Nicolaou $^*$ \and 
        Athanasios Papaioannou$^\|$ \and 
        Guoying Zhao$^{1}$ \and 
        Bj{\"o}rn Schuller$^{2}$ \and 
        Irene Kotsia$^{3}$ \and 
        Stefanos Zafeiriou$^{4}$
}

\institute{    
$^\star$dimitrios.kollias15@imperial.ac.uk\\
$^\dagger$panagiotis.tzirakis12@imperial.ac.uk\\
$^*$m.nicolaou@gold.ac.uk\\
$^\|$a.papaioannou11@imperial.ac.uk\\
$^{1}$guoying.zhao@oulu.fi\\
$^{2}$bjoern.schuller@imperial.ac.uk\\
$^{3}$I.Kotsia@mdx.ac.uk\\
$^{4}$s.zafeiriou@imperial.ac.uk\\
\at
Queen’s Gate, London SW7 2AZ, UK\\
\at
$^*$ Department of Computing, Goldsmiths University
of London, London SE14 6NW, U.K\\
\at
$^{1,4}$Center for Machine Vision and Signal Analysis, University of
Oulu, Oulu, Finland\\
\at
$^3$ Department of Computer Science, Middlesex University
of London, London NW4 4BT, U.K\\
}

\date{Accepted: 29 January 2019 }

\maketitle

\begin{abstract}
Automatic understanding of human affect using visual signals is of great importance in everyday human-machine interactions. Appraising human emotional states, behaviors and reactions displayed in real-world settings, can be accomplished using latent continuous dimensions (e.g., the circumplex model of affect). Valence (i.e., how positive or negative is an emotion) and arousal (i.e., power of the activation of the emotion) constitute popular and effective representations for affect. Nevertheless, the majority of collected datasets this far, although containing naturalistic emotional states, have been captured in highly controlled recording conditions. In this paper, we introduce the Aff-Wild benchmark for training and evaluating affect recognition algorithms. We also report on the results of the First Affect-in-the-wild Challenge  (Aff-Wild Challenge) that was recently organized in conjunction with CVPR 2017 on the Aff-Wild database, and was the first ever challenge on the estimation of valence and arousal in-the-wild. Furthermore, we design and extensively train an end-to-end deep neural architecture which performs prediction of continuous emotion dimensions based on visual cues. The proposed deep learning architecture, AffWildNet, includes convolutional and recurrent neural network (CNN-RNN) layers, exploiting the invariant properties of convolutional features, while also modeling temporal dynamics that arise in human behavior via the recurrent layers. The AffWildNet produced state-of-the-art results on the Aff-Wild Challenge. We then exploit the AffWild database for learning features, which can be used as priors for achieving best performances both for dimensional, as well as categorical emotion recognition, using the RECOLA, AFEW-VA and EmotiW 2017 datasets, compared to all other methods designed for the same goal. The database and emotion recognition models
are available at \url{http://ibug.doc.ic.ac.uk/resources/first-affect-wild-challenge}.


\end{abstract}

\section{Introduction}
\label{intro}
Current research in automatic analysis of facial affect aims at developing systems, such as robots and virtual humans, that will interact with humans in a naturalistic way under real-world settings. To this end, such systems should automatically sense and interpret facial signals relevant to emotions, appraisals and intentions.  Moreover, since real-world settings entail uncontrolled conditions, where subjects operate in a diversity of contexts and environments, systems that perform automatic analysis of human behavior should be robust to video recording conditions, the diversity of contexts and the timing of display. \footnote{It is well known that the interpretation of a facial expression may depend on its dynamics, e.g. posed vs. spontaneous expressions \cite{zeng2009survey}.}  

\begin{figure*}
\centering
\adjincludegraphics[height=9.3cm,width=13.3cm]{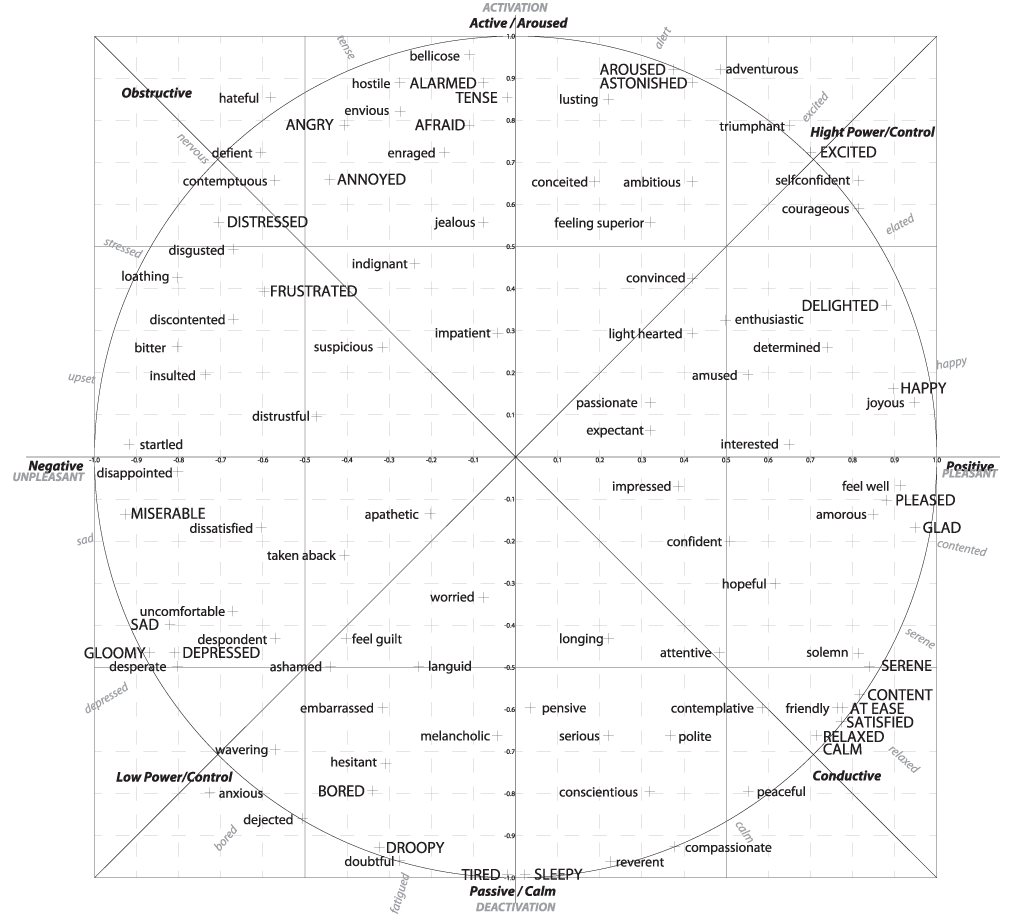} 
\caption{The 2-D Emotion Wheel
}
\label{whicsel}
\end{figure*}

For the past twenty years research in automatic analysis of facial behavior was mainly limited to posed behavior which was captured in highly controlled recording conditions \cite{pantic2005web,valstar2010induced,tian2001recognizing,lucey2010extended}. Some representative datasets, which are still used in many recent works \cite{jung2015joint}, are the Cohn-Kanade database \cite{tian2001recognizing,lucey2010extended},  MMI database  \cite{pantic2005web,valstar2010induced}, Multi-PIE database \cite{gross2010multi} and the BU-3D and BU-4D databases \cite{yin20063d,yin2008high}. 

Nevertheless, it is now accepted by the community that the facial expressions of naturalistic behaviors can be radically different from the posed ones \cite{corneanu2016survey,sariyanidi2015automatic,zeng2009survey}.  Hence, efforts have been made in order to collect subjects displaying naturalistic behavior. Examples include the recently collected EmoPain \cite{Emopain} and UNBC-McMaster \cite{lucey2011painful} databases for analysis of pain, the RU-FACS database of subjects participating in a false opinion scenario \cite{bartlett2006fully} and the SEMAINE corpus\cite{mckeown2012semaine}  which contains recordings of subjects interacting with a Sensitive Artificial Listener (SAL) in controlled conditions. All the above databases have been captured in well-controlled recording conditions and mainly under a strictly defined scenario eliciting  pain.

Representing human emotions has been a basic topic of research in 
psychology. The most frequently used emotion representation 
is the categorical one, including the seven basic categories, i.e., 
Anger, Disgust, Fear, Happiness, Sadness, Surprise and Neutral \cite{dalgleish2000handbook}\cite{cowie2003describing}. 
It is, however, the dimensional emotion representation \cite{whissel1989dictionary},\cite{russell1978evidence} which is more appropriate to represent subtle, i.e., not only extreme, 
emotions appearing in everyday human computer interactions. 
To this end, the 2-D valence and arousal space is the most usual dimensional 
emotion representation. Figure \ref{whicsel}  shows the 2-D Emotion Wheel \cite{plutchik1980emotion},
with valence ranging from very positive to very negative and arousal ranging
from very active to very passive.

Some emotion recognition databases exist in the literature that utilize dimensional emotion representation. 
Examples are the SAL \cite{douglas2008sensitive}, SEMAINE \cite{mckeown2012semaine}, MAHNOB-HCI \cite{soleymani2012multimodal}, Belfast naturalistic \footnote{\label{note1}https://belfast-naturalistic-db.sspnet.eu/}, Belfast induced \cite{sneddon2012belfast}, DEAP \cite{koelstra2012deap}, RECOLA \cite{ringeval2013introducing}, SEWA \footnote{\label{note2}http://sewaproject.eu} and AFEW-VA \cite{kossaifi2017afew} databases. 

Currently, there are many challenges (competitions) in the behavior analysis domain. One such example is the Audio/Visual Emotion Challenges (AVEC) series \cite{valstar2013avec,valstar2014avec,ringeval2015avec,valstar2016avec,ringeval2017avec} which started  in 2011.
The first challenge \cite{schuller2011avec} (2011) used the SEMAINE database for classification purposes by binarizing its continuous values, while the second challenge \cite{schuller2012avec} (2012) used the same database but with its original values. The last challenge (2017) \cite{ringeval2017avec} utilized the SEWA database.  Before this and for two consecutive years (2015 \cite{ringeval2015avec}, 2016 \cite{valstar2016avec}) the RECOLA dataset was used.

However these databases have some of the below  
limitations, as shown in Table \ref{dbs}:
\begin{itemize}
\item[(1)] they contain data recorded in laboratory or controlled
environments.
\item[(2)] their diversity is limited due to the small total number of subjects they contain, the limited amount of head pose variations and present occlusion, the static background or uniform illumination
\item[(3)] the total duration of their included videos is rather short
\end{itemize}

\begin{table}[h]
\caption{Databases annotated for both valence and arousal \& their attributes.}
\label{dbs}
\centering
\scalebox{0.9}{
\begin{tabular}{|c|c|c|c|c|}
\hline
Database & \begin{tabular}{@{}c@{}}no of \\ subjects \end{tabular} & \begin{tabular}{@{}c@{}}no of\\ videos \end{tabular} & \begin{tabular}{@{}c@{}}duration \\of each video \end{tabular}& condition\\
\hline
\begin{tabular}{@{}c@{}}MAHNOB-\\HCI \cite{soleymani2012multimodal} \end{tabular}  & $27$  & $20$ & \begin{tabular}{@{}c@{}} $34.9 - 117$ secs \end{tabular} & controlled \\
\hline
DEAP \cite{koelstra2012deap} & $32$ & $40$  & $1$ min & controlled   \\
\hline
AFEW-VA \cite{kossaifi2017afew} & $<600$  & $600$ & \begin{tabular}{@{}c@{}} $0.5 - 4$ secs \end{tabular} & in-the-wild \\
\hline
SAL \cite{douglas2008sensitive} & $4$ & $24$   & $25$ mins  & controlled \\
\hline
SEMAINE \cite{mckeown2012semaine} & $150$ &  $959$  & $5$ mins & controlled\\
\hline
\begin{tabular}{@{}c@{}}Belfast \\naturalistic \footnoteref{note1}  \end{tabular} & $125$ & $298$  & \begin{tabular}{@{}c@{}} $10 - 60$ secs \end{tabular}  & controlled \\
\hline
\begin{tabular}{@{}c@{}}Belfast\\ induced \cite{sneddon2012belfast} \end{tabular} & $37$ & $37$  & \begin{tabular}{@{}c@{}} $5 - 30$ secs \end{tabular}  & controlled \\
\hline
RECOLA \cite{ringeval2013introducing} & $46$ & $46$   & $5$ mins & controlled\\
\hline
SEWA \footnoteref{note2} & $<398$ & $538$   & $10-30$ secs & in-the-wild\\
\hline
\end{tabular}}

\end{table}

To tackle the aforementioned limitations, we collected the first, to the best of our knowledge, large scale  captured in-the-wild database and annotated it in terms of valence and arousal. To do so, we capitalized on the abundance of data available in video-sharing websites, such as YouTube\cite{youtube2011youtube}\footnote{The collection has been conducted under the scrutiny and approval of the Imperial College Ethical Committee (ICREC). The majority of the chosen videos were under Creative Commons License (CCL). For those videos that were not under CCL, we have contacted the person who created them and asked for their approval to be used in this research.}
and selected videos that display the affective behavior of people,
for example videos that display the behavior of people
when watching a trailer, a movie, a disturbing clip,
or reactions to pranks.  

To this end we have collected 298 videos displaying reactions of 200 subjects, with a total video duration of more than 30 hours. This database has been annotated by 8
lay experts with regards to two continuous emotion dimensions,
i.e. valence and arousal. 
We then organized the Aff-Wild Challenge based on the Aff-Wild database \cite{zafeiriou2017aff}\cite{zafeiriou}, in conjunction with International Conference on Computer Vision \& Pattern Recognition (CVPR) 2017. The participating teams submitted their results to the challenge, outperforming the provided baseline. However, as described later in this paper, the achieved performances were rather low.

For this reason, we capitalized on the Aff-Wild database to build CNN and CNN plus RNN architectures shown to achieve excellent performance on this database, outperforming all previous participants' performances. We have made extensive experimentations, testing structures for combining convolutional and recurrent neural networks and training them altogether as an end-to-end architecture. We have used a loss function that is based on the Concordance Correlation Coefficient (CCC), which we also compare it with the usual Mean Squared Error (MSE) criterion. Additionally, we appropriately fused, within the network structures, two types of inputs, the 2-D facial images - presented at the input of the end-to-end architecture - and the 2-D facial landmark positions - presented at the 1st fully connected layer of the architecture. 

We have also investigated the use of the created CNN-RNN architecture for valence and arousal estimation in other datasets, focusing on the RECOLA and the AFEW-VA ones. Last but not least, taking into consideration the large in-the-wild nature of this database, we show that our network can be also used for other emotion recognition tasks, such as classification of the universal expressions.

The only challenge, apart from last AVEC (2017) \cite{ringeval2017avec}, using 'in-the-wild' data is the series of EmotiW \cite{dhall2013emotion,dhall2014emotion,dhall2015video,dhall2016emotiw,dhall2017individual}. It uses the AFEW dataset, whose samples come from movies, TV shows and series. 
To the best of our knowledge, this is the first time that a dimensional database and features extracted from it, are used as priors for categorical emotion recognition in-the-wild, exploiting the EmotiW Challenge dataset.

\begin{table*}[h]
\caption{Current databases used for emotion recognition in this paper, their attributes and limitations compared to Aff-Wild.}
\label{existing_dbs_limitations}
\centering
\begin{tabular}{|c|c|c|c|c|c|c|}
\hline
Database & model of affect & condition & \begin{tabular}{@{}c@{}}total \\no of frames \end{tabular} & \begin{tabular}{@{}c@{}}no of\\ videos  \end{tabular} & \begin{tabular}{@{}c@{}} no of \\ annotators \end{tabular} & limitations/comments \\
\hline
RECOLA  & \begin{tabular}{@{}c@{}} valence-arousal \\ (continuous) \end{tabular} & controlled & $345,000$ & $46$   & $6$ &\begin{tabular}{@{}c@{}} - laboratory environment \\  - moderate total amount of frames \\ - small number of subjects (46) \end{tabular}  \\
\hline
AFEW  & \begin{tabular}{@{}c@{}} seven basic \\ facial expressions \end{tabular} & in-the-wild & $113,355$ & $1809$   & $3$ & \begin{tabular}{@{}c@{}} - only 7 basic expressions  \\ - small total amount of frames \\ - small number of annotators \\ - imbalanced expression categories   \end{tabular}  \\
\hline
AFEW-VA  & \begin{tabular}{@{}c@{}} valence-arousal \\ (discrete) \end{tabular}  &  in-the-wild & $30,050$ & $600$  & $2$ & \begin{tabular}{@{}c@{}} - very small total amount of frames \\ - discrete valence and arousal values \\ - small number of annotators \end{tabular}   \\
\hline
Aff-Wild & \begin{tabular}{@{}c@{}} valence-arousal \\ (continuous) \end{tabular}& in-the-wild   & $1,224,100$ & $298$ & $8$ & - \\
\hline
\end{tabular}
\end{table*}

To summarize, there exist several databases  for dimensional emotion recognition. However, they have limitations, mostly due to the fact that they are not captured in-the-wild (i.e., not in uncontrolled conditions). This urged us to create the benchmark Aff-Wild database and  organize the Aff-Wild Challenge. The results acquired are presented later in full detail. We proceeded in conducting experiments and building CNN and CNN plus RNN architectures, including the AffWildNet,  producing state-of-the-art results.

The main contributions of the paper are the following:
\begin{description}[leftmargin=!,font=$\bullet$]
\item It is the first time that a large in-the-wild database - with a big variety of: (1) emotional states, (2) rapid emotional changes, (3) ethnicities, (4) head poses, (5) illumination conditions and (6) occlusions - has been generated and used for emotion recognition. 
\item An appropriate state-of-the-art deep neural network (DNN) (AffWildNet) has been developed, which is capable of learning to model all these phenomena. This has not been technically straightforward, as can be verified by comparing the AffWildNet's performance to the performances of other DNNs developed by other research groups which participated in the Aff-Wild Challenge.
\item It is shown that the AffWildNet has been capable of generalizing its knowledge in other emotion recognition datasets and contexts. By learning complex and emotionally rich features of the AffWild, the AffWildNet constitutes a robust prior for both dimensional and categorical emotion recognition. To the best of our knowledge, it is the first time that state-of-the-art performances are achieved in this way.
\end{description}

The rest of the paper is organized as follows. Section 2 presents the databases generated and used in the presented experiments. Section 3 describes the pre-processing and annotation methodologies that we used. Section 4 begins by describing the Aff-Wild Challenge that was organized, the baseline method, the methodologies of the participating teams and their results. It then presents the end-to-end DNNs which we developed and the best performing AffWildNet architecture. Finally experimental studies and results  are presented and discussed, illustrating the above developments. Section 5 describes how the AffWildNet can be used as a prior for other, both dimensional and categorical, emotion recognition problems yielding state-of-the-art results. Finally, Section 6 presents the conclusions and future work following the reported developments.

\section{Existing Databases}
\label{sec:1}

We briefly present the RECOLA, AFEW, AFEW-VA data-bases used for emotion recognition and mention their limitations which lead to the creation of the Aff-Wild database. Table \ref{existing_dbs_limitations} summarizes these limitations, also showing the superior properties of Aff-Wild.

\subsection{RECOLA Dataset}

The REmote
COLlaborative and Affective (RECOLA) data-
base  was introduced by Ringeval et al. \cite{ringeval2013introducing} and it contains natural and spontaneous emotions in the continuous domain (arousal and valence). The corpus includes four modalities: audio, visual, electro-dermal activity and electro-cardiogram. It consists of 46 French speaking subjects being recorded for 9.5\,h recordings in total. The recordings  were annotated for 5\,minutes each by 6 French-speaking annotators (three male, three female). The dataset is divided into three parts, namely, training (16 subjects), validation (15 subjects) and test (15 subjects), in such a way that the gender, age and mother tongue are stratified (i.e., balanced).

The main limitations of this dataset include the tightly controlled laboratory environment, as well as the small number of subjects. It should be also noted that it contains a moderate total number of frames.

\begin{figure*}[h!]
\centering
\adjincludegraphics[height=4.5cm,width=13cm]{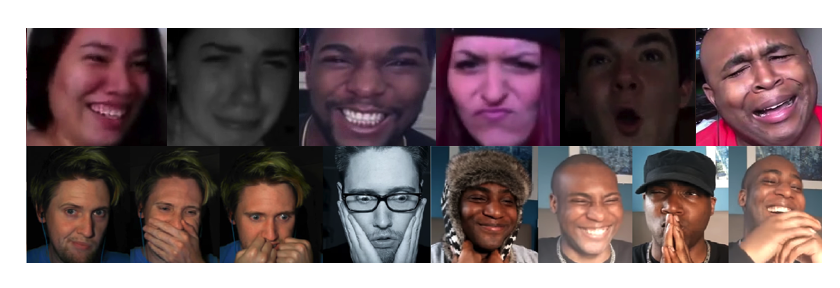} 
\caption{Frames from the Aff-Wild database which show subjects in different emotional states, of different ethnicities, in a variety of head poses, illumination conditions and  occlusions.} 
\label{c123}
\end{figure*}
 
\subsection{The AFEW Dataset}

The series of EmotiW challenges \cite{dhall2013emotion,dhall2014emotion,dhall2015video,dhall2016emotiw,dhall2017individual} make use of the data from the Acted Facial Expression In The Wild (AFEW) dataset \cite{dhall2017individual}. This dataset is a dynamic temporal facial expressions data corpus consisting of close to real world scenes extracted from  movies and reality TV shows. In total it contains 1809 videos. The whole dataset is split into three sets: training set (773 video clips), validation set (383 video clips) and test set (653 video clips). It should be emphasized that both training and validation sets are mainly composed of real movie records, however 114 out of 653 video clips in the test set are real TV clips, thus  increasing the difficulty of the challenge. The number of subjects is more than 330, aged 1-77 years. The annotation is according to 7 facial expressions (Anger, Disgust, Fear, Happiness, Neutral, Sadness and Surprise) and is performed by three annotators. The EmotiW challenges focus on audiovisual classification of each clip into the seven basic emotion categories.

The limitations of the AFEW dataset include its small size (in terms of total number of frames) and its restriction to only seven emotion categories, some of which (fear, disgust, surprise) include a small number of samples.

\subsection{The AFEW-VA Database}

Very recently, a part of the AFEW dataset of the series of EmotiW challenges has been annotated in terms of valence and arousal, thus creating the so called AFEW-VA \cite{kossaifi2017afew} data-
base. In total, it contains 600 video clips that were extracted from feature films and simulate real-world conditions, i.e., occlusions, different illumination conditions and free movements from subjects. The videos range from short (around 10 frames) to longer clips (more than 120 frames). This database includes per-frame annotations of valence and arou-sal.
In total, more than 30,000 frames were annotated for dimensional affect prediction of arousal and valence, using discrete values in the range of [$-10 $, $+10 $]. 

The database's limitations include its small size (in terms of total number of frames), the small number of annotators (only 2) and the use of discrete values for valence and arousal. It should be noted that the 2-D Emotion Wheel (Figure \ref{whicsel}) is a continuous space. Therefore, using discrete only values for valence and arousal provides a rather coarse approximation of the behavior of persons in their everyday interactions. On the other hand, using continuous values can provide improved modeling of the expressiveness and richness of emotional states met in everyday human behaviors.

\subsection{The Aff-Wild Database}

We created a database consisting of 298 videos, with a total length of more than 30 hours. The aim was to collect spontaneous facial behaviors in arbitrary recording conditions. To this end, the videos were collected using the Youtube video sharing web-site. The main keyword that was used to retrieve the videos was "reaction". The database  displays subjects reacting to a variety of stimuli, e.g. viewing an unexpected plot twist of a movie or series, a trailer of a highly anticipated movie, or tasting something hot or disgusting. The subjects display both positive or negative emotions (or combinations of them). In other cases, subjects display emotions while performing an activity (e.g., riding a rolling coaster). In some videos, subjects react on a practical joke, or on positive surprises (e.g., a gift). The
videos contain subjects from different genders and ethnicities with high variations in head pose and lightning.

Most of the videos are in YUV 4:2:0 format, with some of them being in AVI format. Eight subjects have annotated the videos following a methodology similar to the one proposed in \cite{cowie2000feeltrace}, in terms of valence and arousal. An online annotation procedure was used, according to which annotators were watching each video and provided their annotations through a joystick. Valence and arousal range continuously in [$-1 $, $+1 $]. All subjects present in each video have been annotated. The total number of subjects is 200, with 130 of them being male and 70 of them female. 
Table \ref{s_attributes1} shows the general attributes of the Aff-Wild database. Figure \ref{c123} shows some frames from the Aff-Wild database, with people from different ethnicities displaying various emotions, with different head poses and illumination conditions, as well as occlusions in the facial area.

\begin{table}[h]
\caption{Attributes of the Aff-Wild Database}
\label{s_attributes1}
\centering
\begin{tabular}{|c|c|}
\hline
Attribute & Description\\
\hline
Length of videos & $0.10-14.47$ min \\
\hline
Video format & AVI , MP$4$\\
\hline
Average Image Resolution (AIR) & $607 \times 359$\\
\hline
Standard deviation of AIR & $85 \times 11$\\
\hline
Median Image Resolution & $640 \times 360$\\
\hline
\end{tabular}
\end{table}

\begin{figure*}
\centering
 \adjincludegraphics[height=6.5cm,width=17.9cm]{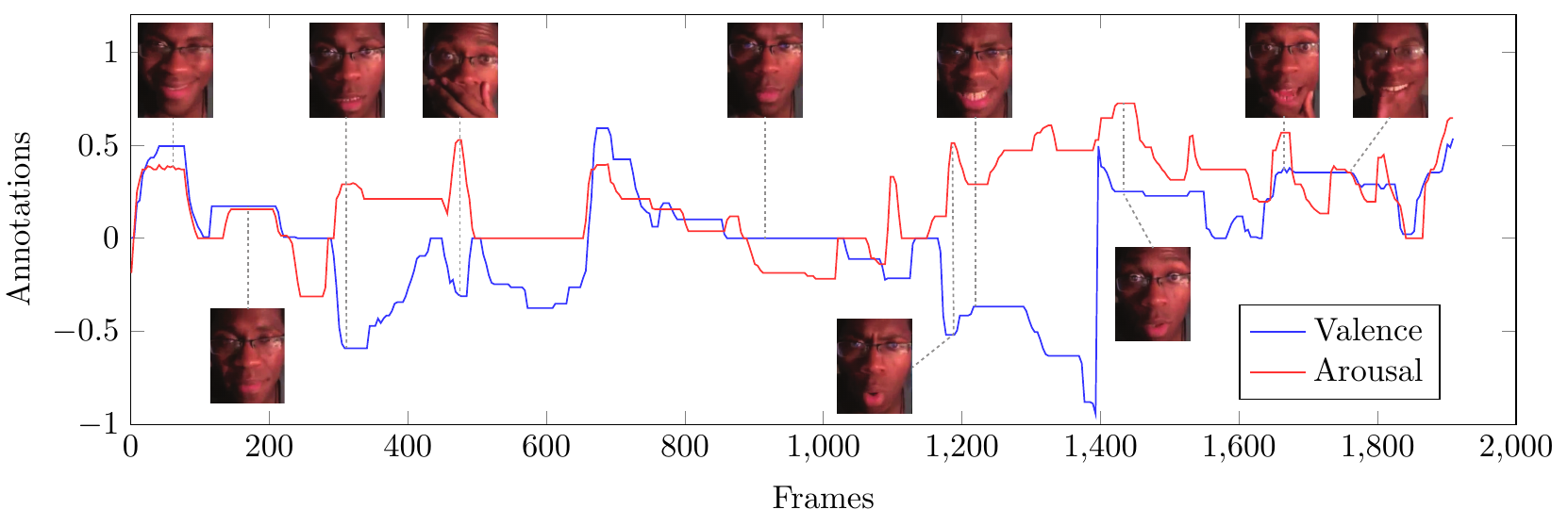} 
\caption{Valence and arousal annotations over a part of a video, along with corresponding frames; illustrating (i) the in-the-wild nature of Aff-Wild (different emotional states, rapid emotional changes, occlusions) and (ii) the use of continuous values for valence and arousal}
\label{trajectory}
\end{figure*}

\begin{figure*}[t]
\centering
 \adjincludegraphics[height=4.7cm,width=15cm,trim={.11\totalheight} {.34\totalheight} {.12\totalheight} {.35\totalheight}] {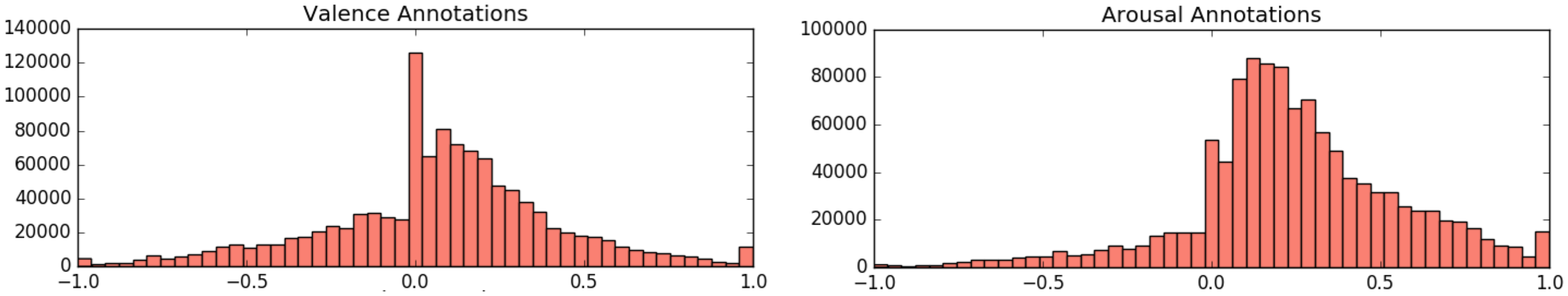} 
\caption{Histogram of valence and arousal annotations of the Aff-Wild database.}
\label{hist}
\end{figure*}

Figure \ref{trajectory} shows an example of annotated valence and arousal values over a part of a video in the Aff-Wild, together with corresponding frames. This illustrates the in-the-wild nature of our database, namely, including many different emotional states, rapid emotional changes and occlusions in the facial areas. Figure \ref{trajectory} also shows the use of continuous values for valence and arousal annotation, which gives the ability to effectively model all these different phenomena.
Figure \ref{hist} provides a histogram for the annotated values for valence and arousal in the generated database.

\section{Data Pre-processing and Annotation}\label{section3}

In this section we describe the pre-processing process of the Aff-Wild videos so as to perform face and facial landmark detection.
Then we present the annotation procedure including:
\begin{itemize}
\item[(1)] Creation of the annotation tool.
\item[(2)] Generation of guidelines for six experts to follow in order to perform the annotation.
\item[(3)] Post-processing annotation: the six annotators watched all videos again, checked their annotations and performed any corrections; two new annotators watched all videos and selected 2-4 annotations that best described each video; final annotations are the mean of the selected annotations by these two new annotators.
\end{itemize}

The detected faces and facial landmarks, as well as the generated annotations are publicly available with the Aff-Wild database.

Finally, we present a statistical analysis of the annotations created for each video, illustrating the consistency of annotations achieved by using the above procedure.

\subsection{Aff-Wild video pre-processing}

VirtualDub \cite{lee2002welcome} was used first so as to trim the raw YouTube videos, mainly at their beginning and end-points, in order to remove useless content (e.g., advertisements). Then, we extracted a total of 1,224,100 video frames using the Menpo software \cite{menpo14}.
In each frame, we detected the faces and generated corresponding bounding boxes, using the method described in  \cite{mathias2014face}. Next, we extracted facial landmarks in all frames using the best performing method as indicated in \cite{chrysos2018comprehensive}. 

During this process, we removed frames in which the bounding box or landmark detection failed. Failures occurred when either the bounding boxes, or landmarks, were wrongly detected, or were not detected at all. The former case was semi-automatically discovered by: (i) detecting significant shifts in the bounding box and landmark positions between consecutive frames and (ii) having the annotators verify the wrong detection in the frames.

\subsection{Annotation tool}

For data annotation, we developed our own application that builds on other existing ones, 
like Feeltrace \cite{cowie2000feeltrace} and Gtrace \cite{cowie2012tracing}.
A time-continuous annotation is performed for each affective dimension, with the annotation process being as follows:
\begin{itemize}
\item[(a)] the user logs in to the application using an identifier (e.g. his/her name) and selects an appropriate joystick; 
\item[(b)] a scrolling list of all videos appears and the user selects a video to annotate;
\item[(c)] a screen appears that shows the selected video and a slider of valence or arousal values ranging in $[-1,1]$;
\item[(d)] the user annotates the video by moving the joystick either up or down;
\item[(e)] finally, a file is created including the annotation values and the corresponding time instances that the annotations are generated.
\end{itemize}

It should be mentioned that the time instances generated in the above step (e), did not generally match the video frame rate. To tackle this problem, we modified/re-sampled the annotation time instances using nearest neighbor interpolation.

\begin{figure}[h]
\begin{tabularx}{\linewidth}{>{\hsize=1.2\hsize}X
                             >{\hsize=1.2\hsize}X}
        \includegraphics[width=\hsize,valign=t]{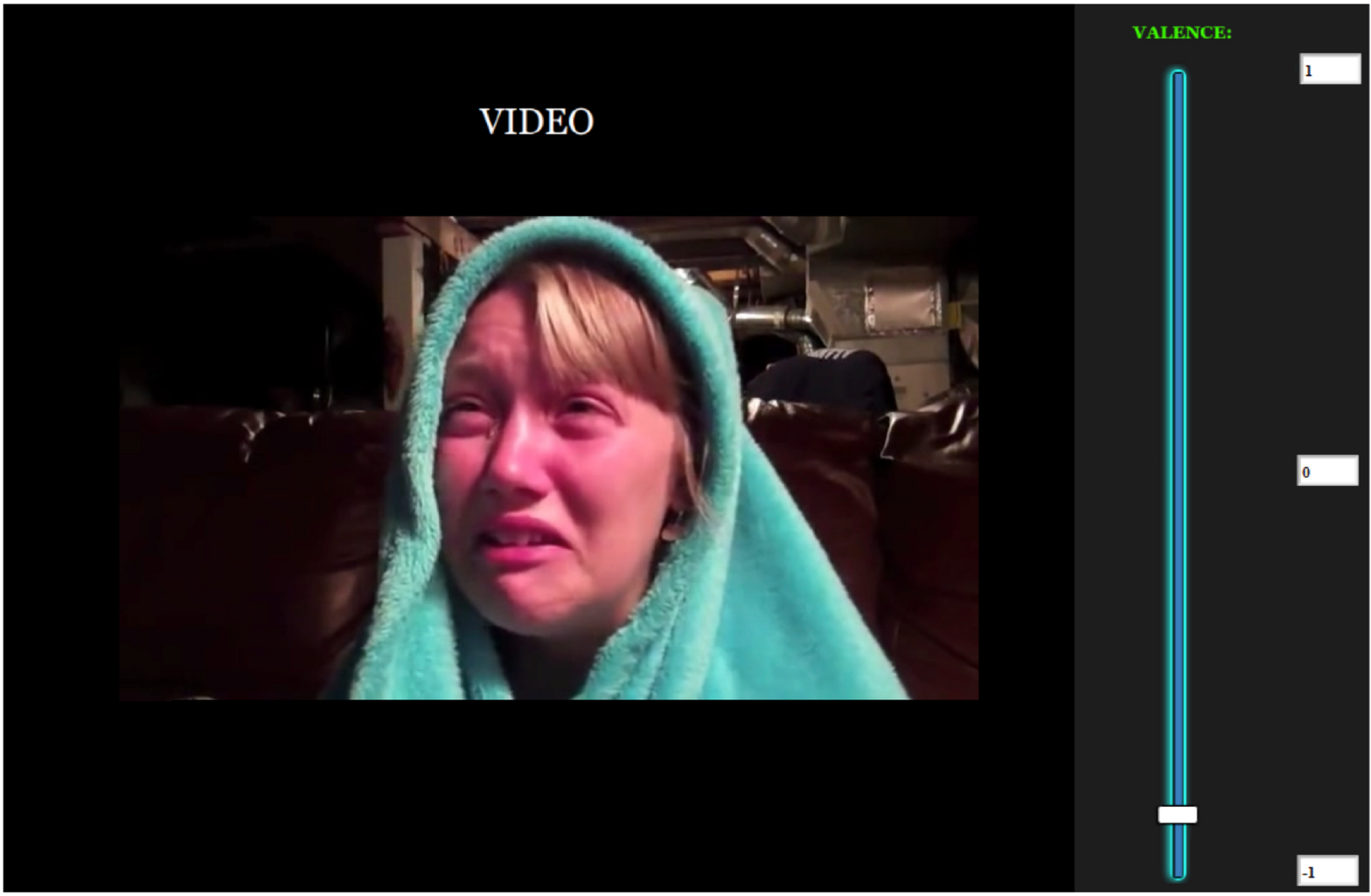}
  \caption{The GUI of the annotation tool when annotating valence (the GUI for arousal is exactly the same).}
  \label{tool_v}
\end{tabularx}
\end{figure}

Figure \ref{tool_v} shows the graphical interface of our tool when annotating valence (the interface for arousal is similar); this corresponds to step (c) of the above described annotation process.

It should also be added that the annotation tool has also the ability to show the inserted valence and arousal annotation while displaying a respective video. This is used for annotation verification in a post-processing step.

\subsection{Annotation guidelines}

Six experts were chosen to perform the annotation task.
Each annotator was instructed orally and through a multi-page document on the procedure to follow for the task. This document included a list of some well identified emotional cues for both arousal and valence, providing a common basis for the annotation task. On top of that the experts used their own appraisal of the subject's emotional state for creating the annotations.\footnote{All annotators were computer scientists who were working on face analysis problems and all had a working understanding of facial expressions.} Before starting the annotation of each video, the experts watched the whole video so as to know what to expect regarding the emotions being displayed in the video.

\subsection{Annotation Post-processing} \label{post_proc}

A post-processing annotation verification step was also performed. 
Every expert-annotator watched  all videos for a second time in order to verify that the recorded annotations were in accordance with the shown emotions in the videos or change the annotations accordingly. In this way, a further validation of annotations was achieved.

After the annotations have been validated by the annotators, a final annotation selection step followed. Two new experts watched all videos and, for every video, selected the annotations (between two and four) which best described the displayed emotions. The mean of these selected annotations constitute the final Aff-Wild labels.

This step is significant for obtaining highly correlated annotations, as shown by the statistical analysis presented next. 

\begin{figure*}[bt]
\centering
\begin{minipage}{.49\textwidth}
\centering
\adjincludegraphics[height=5cm,width=4.5cm,trim={.28\width} {.0\totalheight} {.23\width} {.0\totalheight}]{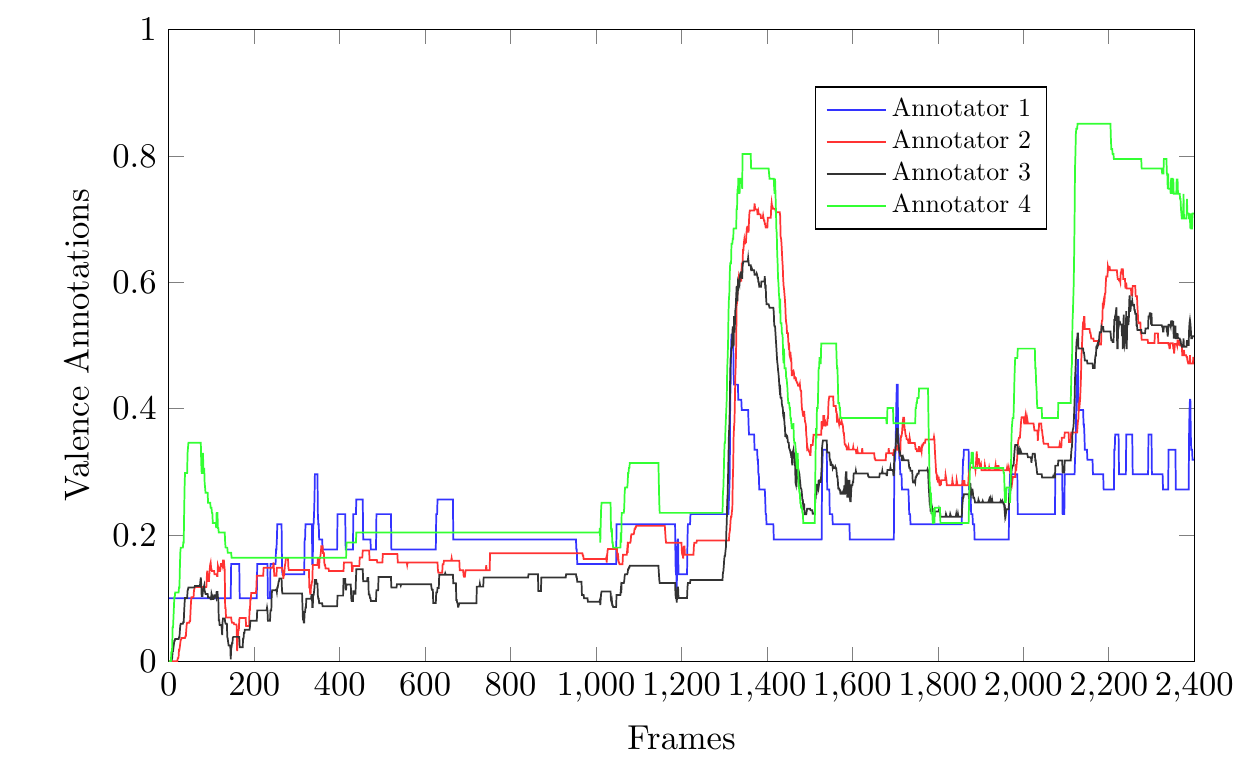} \\
\qquad(a)
\label{vall}
\end{minipage}
\begin{minipage}{.49\textwidth}
\centering
\adjincludegraphics[height=5cm,width=4.5cm,trim={.26\width} {.0\totalheight} {.25\width} {.0\totalheight}]{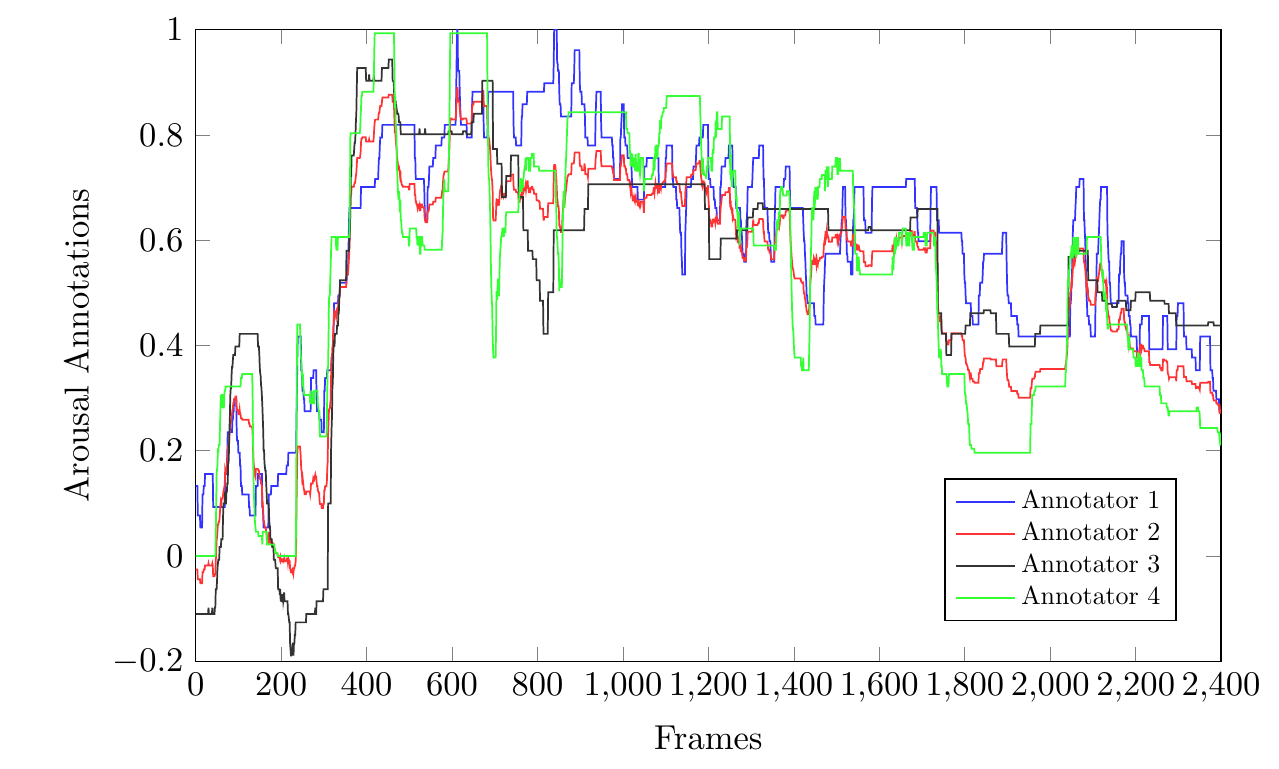}\\
\qquad\qquad(b)
\label{arr}
\end{minipage}
\caption{The four selected annotations in a video segment for (a) valence and (b) arousal. In both cases, the value of MAC-S (mean of average correlations between these four annotations) is 0.70. This value is similar to the mean MAC-S obtained over all Aff-Wild.}
\label{val_ar_0_6}
\end{figure*}

\begin{figure*}[bt]
\centering
\begin{minipage}{.49\textwidth}
\centering
\adjincludegraphics[height=5cm,width=4.2cm,trim={.3\width} {.0\totalheight} {.23\width} {.0\totalheight}]{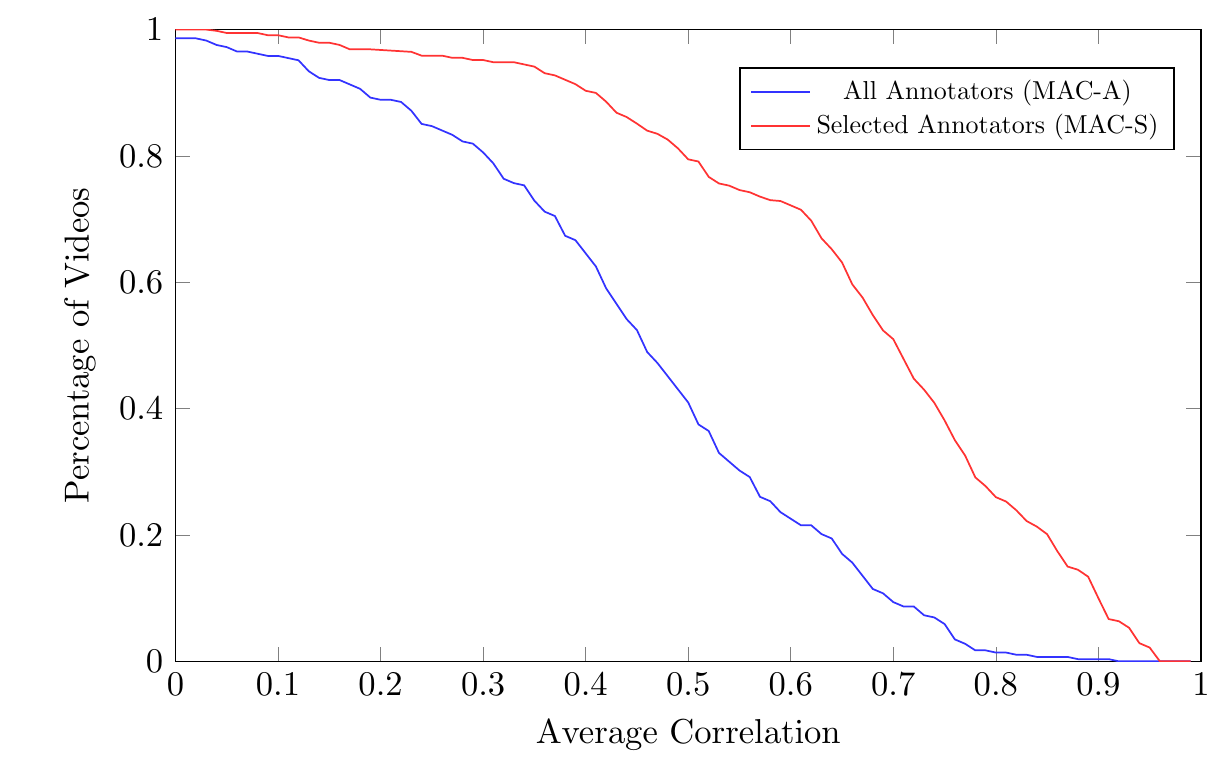} \\
\qquad(a)
\label{vall_mac}
\end{minipage}
\begin{minipage}{.49\textwidth}
\centering
\adjincludegraphics[height=5cm,width=4.2cm,trim={.27\width} {.0\totalheight} {.26\width} {.0\totalheight}]{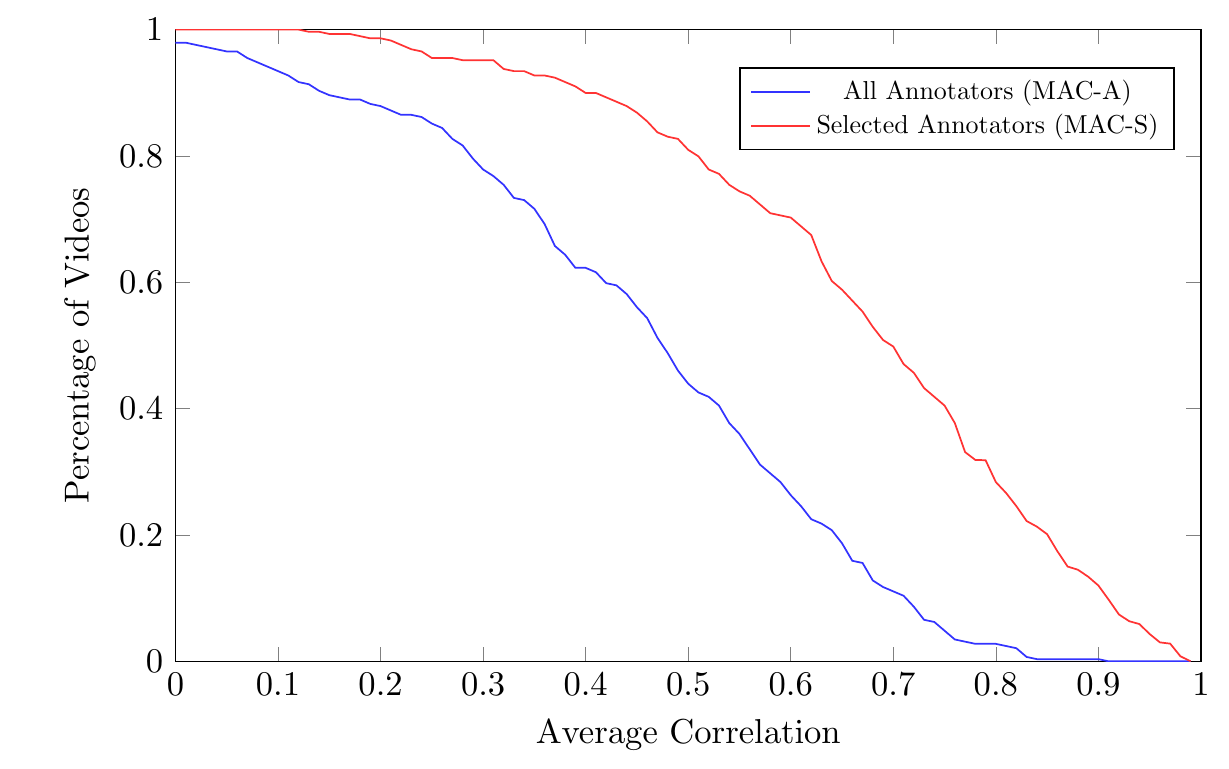}\\
\qquad\qquad(b)
\label{arr_mac}
\end{minipage}
\caption{The cumulative distribution of MAC-S (mean of average inter-selected-annotator correlations) and MAC-A (mean of average inter-annotator correlations) values over all Aff-Wild videos for valence (Figure \ref{mean_corr_perc}a) and arousal (Figure \ref{mean_corr_perc}b). The Figure shows the percentage of videos with a MAC-S/MAC-A  value greater or equal to the values shown in the horizontal axis. The mean MAC-S value, corresponding to a value of 0.5 in the vertical axis, is 0.71 for valence and 0.70 for arousal.}
\label{mean_corr_perc}
\end{figure*}

\begin{figure*}[bt]
\centering
\begin{minipage}{.49\textwidth}
\centering
\adjincludegraphics[height=5cm,width=4.2cm,trim={.3\width} {.0\totalheight} {.23\width} {.0\totalheight}]{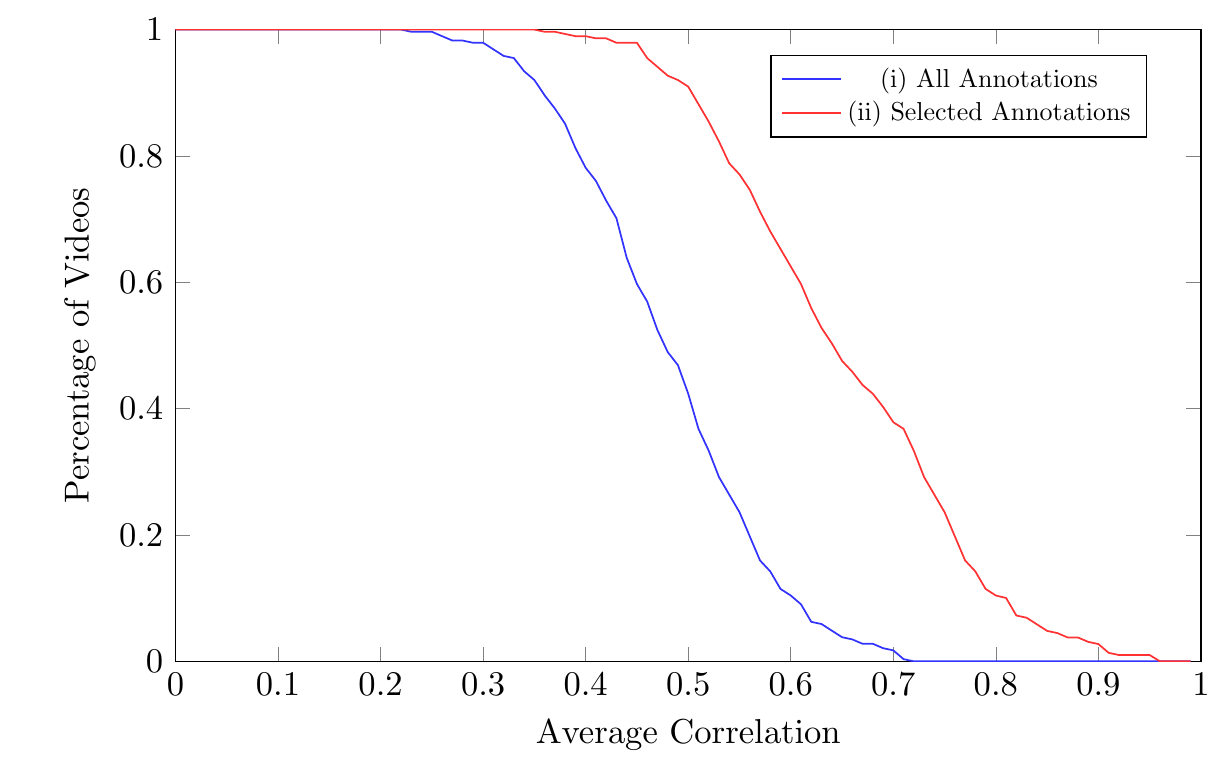} \\
\qquad(a)
\label{vall_mac_lands}
\end{minipage}
\begin{minipage}{.49\textwidth}
\centering
\adjincludegraphics[height=5cm,width=4.2cm,trim={.27\width} {.0\totalheight} {.26\width} {.0\totalheight}]{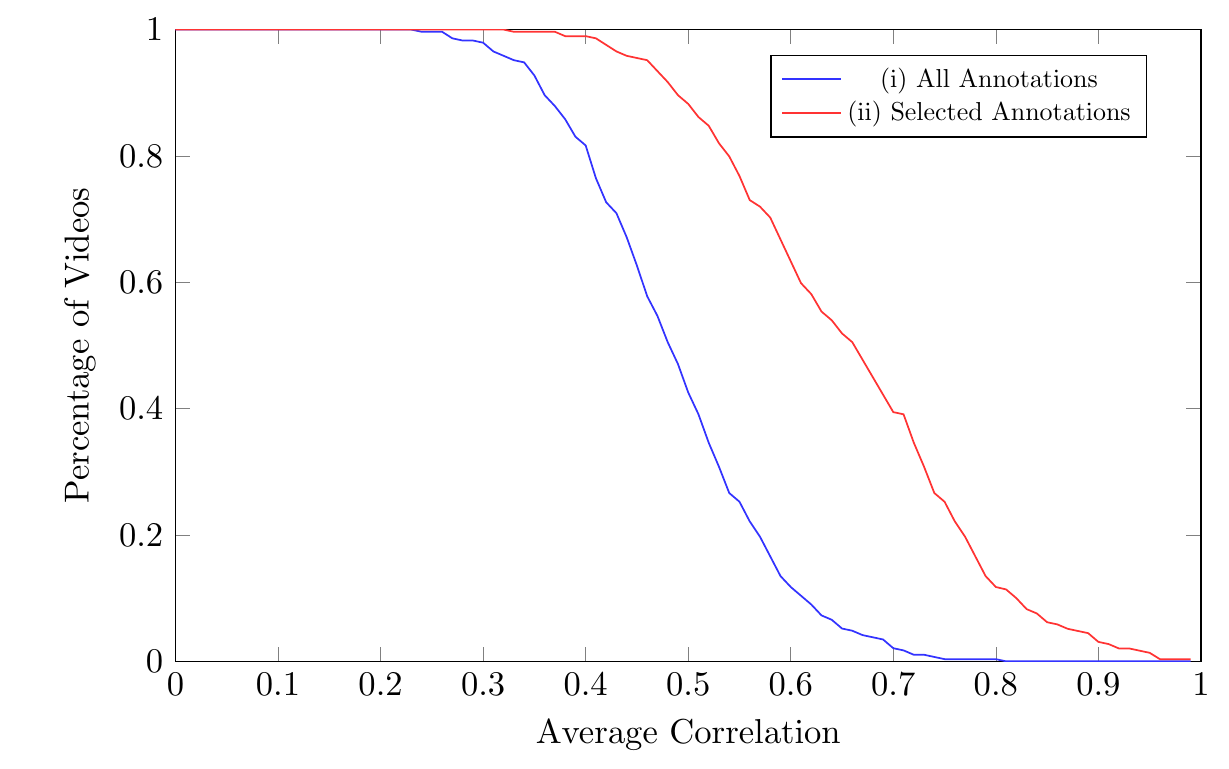}\\
\qquad\qquad(b)
\label{arr_mac_lands}
\end{minipage}
\caption{The cumulative distribution of the correlation between landmarks and the average of (i) all or (ii) selected annotations over all Aff-Wild videos for valence (Figure \ref{mean_corr_perc_lands}a) and arousal (Figure \ref{mean_corr_perc_lands}b). The Figure shows the percentage of videos with a correlation value greater or equal to the values shown in the horizontal axis.}
\label{mean_corr_perc_lands}
\end{figure*}

\subsection{Statistical Analysis of Annotations}

In the following we provide a quantitative and rich statistical analysis of the achieved Aff-Wild labeling. At first, for each video, and independently for valence and arousal, we computed:

\begin{itemize}
\item[(i)] the inter-annotator correlations, i.e., the correlations of each one of the six annotators with all other annotators, which resulted in five correlation values per annotator;
\item[(ii)] for each annotator, his/her average inter-annotator correlations, resulting in one value per annotator; the mean of those six average inter-annotator correlations value is denoted next as MAC-A;
\item[(iii)] the average inter-annotator correlations, across only the selected annotators, as described in the previous subsection, resulting in one value per selected annotator; the mean of those 2-4 average inter-selected-annotator correlations values is denoted next as MAC-S.
\end{itemize}

We then computed over all videos and independently for valence and arousal, the mean of MAC-A and the mean of MAC-S computed in (ii) and (iii) above. The mean MAC-A is  0.47 for valence and 0.46 for arousal, whilst the mean MAC-S for valence is 0.71 and for arousal 0.70. An example set of annotations is shown in Figure \ref{val_ar_0_6}, in an effort to further clarify the obtained  MAC-S values. 
 It shows the four selected annotations in a video segment for valence and arousal, respectively, with MAC-S value of 0.70 (similar to the mean MAC-S value obtained over all Aff-Wild). 

In addition, Figure \ref{mean_corr_perc} shows the cumulative distribution of MAC-S and MAC-A values over all Aff-Wild videos for valence (Figure \ref{mean_corr_perc}a) and arousal (Figure \ref{mean_corr_perc}b). In each case, two curves are shown. Every point $(x,y)$ on these curves has a $y$ value showing the percentage of videos with a (i) MAC-S (red curve) or (ii) MAC-A (blue curve) value greater or equal to $x$; the latter denotes an average correlation in $[0,1]$. It can be observed that the mean MAC-S value, corresponding to a value of 0.5 in the vertical axis, is 0.71 for valence and 0.70 for arousal. These plots also illustrate that the MAC-S values are much higher than the corresponding MAC-A values in both valence and arousal annotation, verifying the effectiveness of the annotation post-processing procedure.

Next, we conducted similar experiments for the valence/ arousal average annotations and the facial landmarks in each video, in order to evaluate the correlation of annotations to landmarks.  To this end, we utilized Canonical Correlation Analysis (CCA)~\cite{eps259225}.
In particular, for each video and independently for valence and arousal, we computed the correlation between landmarks and the average of (i) all or (ii)  selected annotations.

Figure \ref{mean_corr_perc_lands} shows the cumulative distribution of these correlations over all Aff-Wild videos for valence (Figure \ref{mean_corr_perc_lands}a) and arousal (Figure \ref{mean_corr_perc_lands}b), similarly to Figure \ref{mean_corr_perc}. Results of this analysis verify that the annotator-landmark correlation is much higher in the case of selected annotations than in the case of all annotations.

\section{Developing the AffWildNet} \label{sec4}

This section begins by presenting the first Aff-Wild Challenge that was organized based on the Aff-Wild database and held in conjunction with CVPR 2017. It includes short descriptions and results of the algorithms of the six research groups that participated in the challenge. Although the results are promising, there is much room for improvement.

For this reason we developed our own CNN and CNN plus RNN architectures based on the Aff-Wild database. We propose the AffWildNet as the best performing among the developed architectures. Our developments,  ablation studies and discussions are presented next.

\subsection{The Aff-Wild Challenge}

The training data (i.e., videos and annotations) of the Aff-Wild challenge were made publicly available on the 30th of January 2017, followed by the release of the test videos (without annotations). The participants were given the freedom to split the data into train and validation sets, as well as to use any other dataset. The maximum number of submitted entries for each participant was three.
Table \ref{s_attributes} summarizes the specific attributes (numbers of males, females, videos, frames) of the training and test sets of the challenge.

\begin{table}[h]
\caption{Attributes of Training and Test sets of Aff-Wild.}
\label{s_attributes}
\centering
\begin{tabular}{|c|c|c|c|c|}
\hline
Set & \begin{tabular}{@{}c@{}}  no of \\ males \end{tabular} & \begin{tabular}{@{}c@{}} no of \\ females \end{tabular} & \begin{tabular}{@{}c@{}}  no of \\ videos \end{tabular} & \begin{tabular}{@{}c@{}}  total no of \\ frames \end{tabular}\\
\hline
Training & $106$  & $48$ & $252$  & $1,008,650$  \\
\hline
Test & $24$ & $22$ & $46$ & $215,450$ \\
\hline
\end{tabular}
\end{table}

In total, ten different research groups downloaded the Aff-Wild database. Six of them made experiments and submitted their results to the workshop portal. Based on the performance they obtained on the test data, three of them were selected to present their results to the workshop. 

\begin{table*}[h]
\caption{Concordance Correlation Coefficient (CCC) and Mean Squared Error
(MSE) of valence \& arousal predictions provided by the methods of the three participating teams and the baseline architecture. A higher CCC and a lower MSE value indicate a better performance.}
\label{acc}
\parbox{.45\linewidth}{
\centering
\begin{tabular}{ |c||c|c|c| }
 \hline
 \multicolumn{1}{|c||}{Methods} & \multicolumn{3}{c|}{CCC}  \\
 \hline
  & Valence & Arousal & Mean Value \\
 \hline
MM-Net & 0.196 & 0.214   &  0.205    \\
 \hline 
\textbf{FATAUVA-Net} &  \textbf{0.396} & 0.282 &   \textbf{0.339}     \\
 \hline 
DRC-Net &  0.042 & \textbf{0.291} & 0.167    \\
\hline 
Baseline & 0.150 & 0.100   &  0.125    \\
\hline
\end{tabular}
}
\hfil
\parbox{.45\linewidth}{
\centering
\begin{tabular}{ |c||c|c|c| }
 \hline
 \multicolumn{1}{|c||}{Methods} & \multicolumn{3}{c|}{MSE}  \\
 \hline
  & Valence & Arousal & Mean Value\\
 \hline
MM-Net  & 0.134 & \textbf{0.088}  & 0.111    \\
 \hline 
\textbf{FATAUVA-Net}  & \textbf{0.123} & 0.095 & \textbf{0.109}   \\
 \hline 
DRC-Net  & 0.161 & 0.094  & 0.128   \\
\hline 
Baseline   & 0.130 & 0.140  & 0.135   \\
\hline
\end{tabular}
}
\end{table*}

Two criteria were considered for evaluating the performance of the networks. The first one is Concordance Correlation Coefficient (CCC) \cite{lawrence1989concordance}, which is widely used in measuring the performance of dimensional emotion recognition methods, e.g., the series of AVEC challenges. CCC evaluates the agreement between two time series (e.g., all video annotations and predictions) by scaling their correlation coefficient with their mean square difference. In this way, predictions that are well correlated with the annotations but shifted in value are penalized in proportion to the deviation. CCC takes values in the range $[-1,1]$, where $+1$ indicates perfect concordance and $-1$ denotes perfect discordance. The highest the value of the CCC the better the fit between annotations and predictions, and therefore high values are desired.
The mean value of CCC for valence and arousal estimation was adopted as the main evaluation criterion. 
CCC is defined as follows:

\begin{equation} \label{eq:1}
\rho_c = \frac{2 s_{xy}}{s_x^2 + s_y^2 + (\bar{x} - \bar{y})^2} = \frac{2s_x  s_y \rho_{xy}}{s_x^2 + s_y^2 + (\bar{x} - \bar{y})^2},
\end{equation}

\noindent
where $\rho_{xy}$ is the Pearson Correlation Coefficient (Pearson CC), $s_x$ and $s_y$ are the variances of all video valence/arousal annotations and predicted values, respectively and $s_{xy}$ is the corresponding covariance value.

The second criterion is the Mean Squared Error (MSE), which is defined as follows:

\begin{equation} \label{eq:2}
MSE = \frac{1}{N} \sum_{i=1}^{N} (x_i-y_i)^2 ,
\end{equation}

\noindent
where $x$ and $y$ are the (valence/arousal) annotations and predictions, respectively, and $N$ is the total number of samples.
The MSE gives us a rough indication of how the derived emotion model is behaving, providing a simple comparative metric. A small value of MSE is desired.

\subsubsection{Baseline Architecture}

The baseline architecture for the challenge was based on the CNN-M \cite{chatfield2014return} network, as a simple model that could be used to initiate the procedure. In particular, our network used the convolutional and pooling parts of CNN-M having been trained on the FaceValue dataset \cite{albanie16learning}. On top of that we added one 4096-fully connected layer and a 2-fully connected layer that provides the valence and arousal predictions. The interested reader can refer to Appendix A for a short description and the structure of this architecture. 

The input to the network were the facial images resized to resolution of $224 \times 224 \times 3$, or  $96 \times 96 \times 3$, with the intensity values being normalized to the range $[-1, 1]$.

In order to train the network, we utilized the Adam optimizer algorithm; the batch size was set to $80$, and the initial learning rate was set to $0.001$. Training was performed on a single GeForce GTX TITAN X GPU and the training time was about 4-5 days. The platform used for this implementation was Tensorflow \cite{tensorflow2015-whitepaper}.

\subsubsection{Participating Teams' Algorithms}

The three papers accepted to this challenge are briefly reported below, while Table \ref{acc} compares the acquired results (in terms of CCC and MSE) by all three methods and the baseline network. As one can see, FATAUVA-Net \cite{weichi} has provided the best results in terms of the mean CCC and mean MSE for valence and arousal.

We should note that after the end of the challenge, more groups enquired about the Aff-Wild database and sent results for evaluation, but here we report only on the  teams that participated in the challenge.

In the MM-Net method \cite{jianshu}, a variation of a deep convolutional residual neural network (ResNet) \cite{he2016deep} is first presented for affective level estimation of facial expressions. Then, multiple memory networks are used to model temporal relations between the video frames. Finally, ensemble models are used to combine the predictions of the multiple memory networks, showing that the latter steps improve the initially obtained performance, as far as MSE is concerned, by more than 10\%.

In the FATAUVA-Net method \cite{weichi}, a deep learning framework is presented, in which a core layer, an attribute layer, an action unit (AU) layer and a valence-arousal layer are trained sequentially. The core layer is a series of convolutional layers, followed by the attribute layer which extracts facial features. These layers are applied to supervise the learning of AUs. Finally, AUs are employed as mid-level representations to estimate the intensity of valence and arousal. 

In the DRC-Net method \cite{hasani}, three neural network-based methods which are based on Inception-ResNet \cite{szegedy2017inception} modules redesigned specifically for the task of facial affect estimation are presented and compared. These methods are: Shallow Inception-ResNet, Deep Inception-ResNet, and Inception-ResNet with Long Short Term Memory \cite{hochreiter1997long}. Facial features are extracted in different scales and  both, the valence and arousal, are simultaneously estimated in each frame. Best results are obtained by the Deep Inception-ResNet method.

All participants applied deep learning methods to the problem of emotion analysis of the video inputs. The following conclusions can be drawn from the reported results. First, CCC of arousal predictions was really low for all three methods. 
Second, MSE of valence predictions was high for all three methods and CCC was low, except for the winning method. This illustrates the difficulty in recognizing emotion in-the-wild, where, for instance, illumination conditions differ, occlusions are present and different head poses are met.

\subsection{Deep Neural Architectures \& Ablation Studies}\label{ablation_studies}

Here, we present our developments and ablation studies towards designing deep CNN and CNN plus RNN architectures for the Aff-Wild. We present the proposed architecture, AffWildNet, which is a CNN plus RNN network that produced the best results in the database.

\subsubsection{The Roadmap} \label{roadmap}

\begin{itemize}
\item[A.] We  considered two network settings:

\begin{itemize}
\item[(1)] a CNN network trained in an end-to-end manner, i.e., using raw intensity pixels, to produce 2-D predictions of valence and arousal,
\item[(2)] a RNN stacked on top of the CNN to capture temporal information in the data, before predicting the affect dimensions; this was also trained in an end-to-end manner.
\end{itemize}

To extract features from the frames we experimented with three CNN architectures, namely, ResNet-50, VGG-Face \cite{parkhi2015deep} and VGG-16 \cite{simonyan2014very}. To consider the contextual information in the data (RNN case) we experimented with both the Long Short-Term Memory (LSTM) and the Gated Recurrent Unit (GRU) \cite{chung2014empirical} architectures.

\begin{figure*}
\centering
\adjincludegraphics[height=8cm,width=12cm]{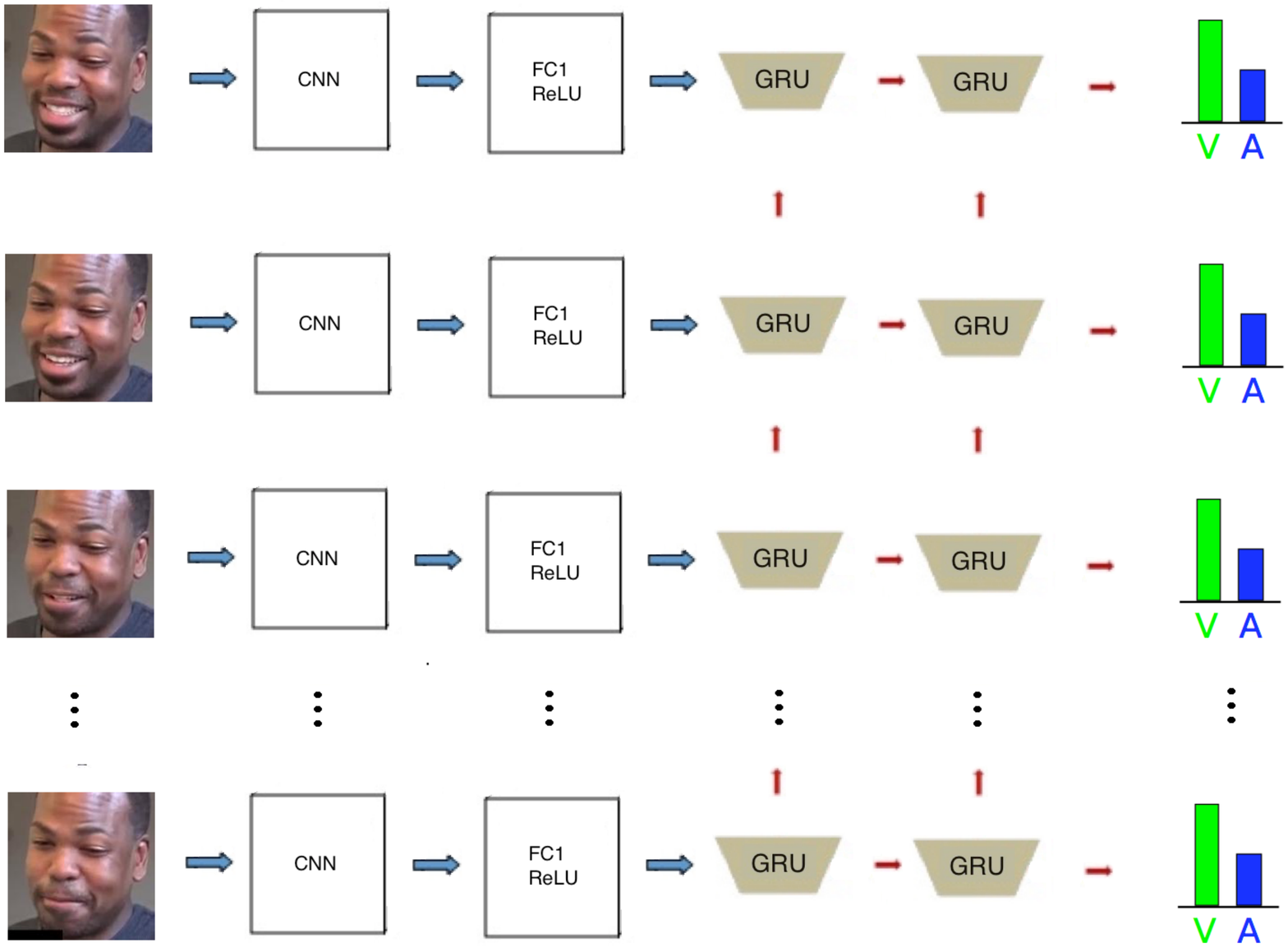} 
\caption{The AffWildNet: it consists of convolutional and pooling layers of either VGG-Face or ResNet-50 structures (denoted as CNN), followed by a fully connected layer (denoted as FC1) and two RNN layers with GRU units (V and A  stand for valence and arousal respectively).}
\label{affwildnet}
\end{figure*}

\mbox{}\\

\item[B.] To further boost the performance of the networks, we also experimented with the use of facial landmarks. Here we should note that the facial landmarks are provided on-the-fly for training and testing the networks. The following two scenarios were tested:

\begin{itemize}
\item[(1)] The networks were applied directly on cropped facial video frames of the generated database.
\item[(2)] The networks were trained on both the facial video frames as well as the facial landmarks corresponding to the same frame.
\end{itemize}

\mbox{}\\

\item[C.] Since the main evaluation criterion of the Aff-Wild Challenge was the mean value of CCC for valence and arousal, our loss function was based on that criterion and was defined as:

\begin{equation} \label{eq:3}
\mathcal{L}_{total} = 1 - \frac{\rho_a + \rho_v}{2},
\end{equation}

\noindent
where $\rho_a$ and $\rho_v$ are the CCC for the arousal and valence, respectively.

\mbox{}\\

\item[D.]  In order to have a more balanced dataset for training, we performed data augmentation, mainly through oversampling by duplicating \cite{more2016survey} some data from the Aff-Wild database. We copied small video parts showing less-populated valence and arousal values. In particular, we duplicated consecutive video frames that had negative valence and arousal values, as well as positive valence and negative arousal values. As a consequence, the training set consisted of about 43\% of positive valence and arousal values, 24\% of negative valence and positive arousal values, 19\% of positive valence and negative arousal values and 14\% of negative valence and arousal values. Our main target has been a trade-off between generating balanced emotion sets and avoiding to severely change the content of videos.

\end{itemize}

\subsubsection{Developing CNN architectures for the Aff-Wild} \label{cnn_dev}

For the CNN architectures, we considered the ResNet-50 and VGG-16 networks, pre-trained on the ImageNet \cite{deng2009imagenet} dataset that has been broadly used for state-of-the-art object detection. We also considered the VGG-Face network, pre-trained for face recognition on the VGG-Face dataset \cite{parkhi2015deep}. The VGG-Face has proven to provide the best results, as reported next in the experimental section. It is worth mentioning that in our experiments we have trained those architectures for predicting both valence and arousal at their output, as well as for predicting valence and arousal separately. The obtained results were similar in the two cases. In all experiments presented next, we focus on the simultaneous prediction of valence and arousal. 

The first architecture we utilized was the deep residual network (ResNet) of 50 layers \cite{he2016deep}, on top of which we stacked a 2-layer fully connected (FC) network. For the first FC layer, best results have been obtained when using 1500 units. For the second FC layer, 256 units provided the best results. An output layer with two linear units followed providing the valence and arousal predictions.
The interested reader can refer to Appendix A for a short description and the structure of this architecture.

The other architecture that we utilized was based on the convolutional and pooling layers of VGG-Face or VGG-16 networks, on top of which we stacked a 2-layer FC network. For the first and second FC layers, best results have been obtained when using 4096 units. An output layer followed, including two linear units, providing the valence and arousal predictions. The interested reader can refer to Appendix A for a short description and the structure of this architecture as well.

In the case when landmarks were used (scenario B.2 in subsection \ref{roadmap}), these were input to the first FC layer along with: i) the outputs of the ResNet-50, or ii) the outputs of the last pooling layer of the VGG-Face/VGG-16. In this way, both outputs and landmarks were mapped to the same feature space before performing the prediction.

With respect to parameter selection in those CNN architectures, we have used a batch size in the range $10 - 100$ and a constant learning rate value in the range $0.00001 - 0.001$. The best results have been obtained with batch size equal to 50 and learning rate equal to $0.0001$. The dropout probability value has been set to $0.5$.

\subsubsection{Developing CNN plus RNN architectures for the Aff-Wild}\label{cnn-rnn_dev}

In order to consider the contextual information in the data, we developed a CNN-RNN architecture, in which the RNN part was fed with the outputs of either the first, or the second fully connected layer of the respective CNN networks.

The structure of the RNN, which we examined, consisted of one or two hidden layers, with $100 - 150$ units, following either the LSTM neuron model with peephole connections, or the GRU neuron model. Using one fully connected layer in the CNN part and two hidden layers in the RNN part, including GRUs, has been found to provide the best results. An output layer followed, including two linear units, providing the valence and arousal predictions.

Table \ref{rnn} shows the configuration of the CNN-RNN architecture. The CNN part of this architecture was based on the convolutional and pooling layers of the CNN architectures described above (VGG-Face, or ResNet-50) that was followed by a fully connected layer.
Note that in the case of scenario B.2 of subsection \ref{roadmap}, both the outputs of the last pooling layer of the CNN, as well as the 68 landmark 2-D positions ($68 \times  2$ values) were provided as inputs to this fully connected layer. Table \ref{rnn} shows the respective number of units for the GRU and the fully connected layers. We call this CNN plus RNN architecture AffWildNet and illustrate it in Figure \ref{affwildnet}.

\begin{table}[h]
\caption{The AffWildNet architecture: the fully connected 1 layer has 4096, or 1500 hidden units, depending on whether VGG-Face or ResNet-50 is used.}
\label{rnn}
\centering
\begin{tabular}{|c|c|c|}
\hline
block 1 & \begin{tabular}{@{}c@{}} VGG-Face or ResNet-50 \\ conv \& pooling parts \end{tabular} & \\
\hline
block 2 &fully connected 1 & 4096 or 1500 \\
&dropout &\\
\hline
block 3 & GRU layer 1 & 128\\
&dropout &\\
\hline
block 4 & GRU layer 2 & 128\\
\hline
block 5 &fully connected 2 &2\\
\hline
\end{tabular}
\end{table}

\begin{table*}[h]
\caption{\looseness-1CCC and MSE based evaluation of valence \& arousal predictions provided by the VGG-Face (using the mean of annotators’ values, or  using only one annotator values; when landmarks were or were not given as input to the network). }
\label{table2}
\parbox{.5\linewidth}{
\centering
\begin{tabular}{ |c||c|c|c|c| }
 \hline
  \multicolumn{1}{|c||}{CCC} & \multicolumn{2}{c|}{\begin{tabular}{@{}c@{}} With \\ Landmarks  \end{tabular}  }  & \multicolumn{2}{c|}{\begin{tabular}{@{}c@{}} Without \\ Landmarks  \end{tabular}} \\
 \hline
    & Valence & Arousal & Valence & Arousal   \\
 \hline
\begin{tabular}{@{}c@{}} One \\ Annotator \end{tabular} & 0.39 & 0.27 & 0.35 & 0.25 \\ 
\hline
\begin{tabular}{@{}c@{}} Mean of \\ Annotators \end{tabular} & \textbf{0.51} & \textbf{0.33} & 0.44 & 0.32 \\ 
 \hline
\end{tabular}
}
\hfill
\parbox{.47\linewidth}{
\centering
\begin{tabular}{ |c||c|c|c|c| }
 \hline
 \multicolumn{1}{|c||}{MSE} & \multicolumn{2}{c|}{\begin{tabular}{@{}c@{}} With \\ Landmarks  \end{tabular}  }  & \multicolumn{2}{c|}{\begin{tabular}{@{}c@{}} Without \\ Landmarks  \end{tabular}} \\
 \hline
    & Valence & Arousal & Valence & Arousal   \\
 \hline
\begin{tabular}{@{}c@{}} One \\ Annotator \end{tabular} & 0.15 & 0.13 & 0.16 & 0.14 \\  
\hline
\begin{tabular}{@{}c@{}} Mean of \\ Annotators \end{tabular} & \textbf{0.10} & \textbf{0.08} & 0.12 & 0.11   \\
 \hline
\end{tabular}
}
\end{table*}

Network evaluation has been performed by testing different parameter values. The parameters included: the batch size and sequence length used for network parameter updating, the value of the learning rate and the dropout probability value. Final selection of these parameters was similar to the CNN cases, apart from the sequence length which was selected in the range $50 - 200$ and batch size that was selected in the range $2 - 10$. Best results have been obtained with sequence length $80$ and batch size $4$. We note that all deep learning architectures have been implemented in the Tensorflow platform.

\subsection{Experimental Results}

In the following we present the affect recognition results obtained when applying the above derived CNN-only and CNN plus RNN architectures to the Aff-Wild database.

At first, we have trained the VGG-Face network using two different annotations. One, which is provided in the Aff-Wild database, is the average of the selected (as described in subsection \ref{post_proc}) annotations.
The second is that of a single annotator (the one with the highest correlation to the landmarks). It should be mentioned that the latter is generally less smooth than the former, average, one. Hence, they are more difficult to be modeled. Then, we tested the two trained networks in two scenarios, as described in subsection \ref{roadmap} case B, using/not using the 68 2-D landmark inputs.

The results are summarized in Table \ref{table2}. As was expected, better results were obtained when the mean of annotations was used. Moreover, Table \ref{table2} shows that there is a notable improvement in the performance, when we also used the 68 2-D landmark positions as input data.

Next, we examined the use of various numbers of hidden layers and hidden units per layer when training and testing the VGG-Face-GRU network. Some characteristic selections and their corresponding performances are shown in Table \ref{table4}. It can be seen that the best results have been obtained when the RNN part of the network consisted of 2 layers, each of 128 hidden units.

\begin{table}[h]
\caption{Obtained CCC values for valence \& arousal estimation, when changing the number of hidden units \& hidden layers in the VGG-Face-GRU architecture. A higher CCC value indicates a better performance.}
\label{table4}
\centering
\begin{tabular}{ |c||c|c|c|c|  }
 \hline
 \multicolumn{1}{|c||}{CCC} & \multicolumn{2}{c|}{1 Hidden Layer} & \multicolumn{2}{c|}{2 Hidden Layers} \\
 \hline
  Hidden Units & Valence & Arousal & Valence & Arousal \\
 \hline
100 & 0.44 & 0.36 & 0.50  & 0.41       \\
 \hline
128 &0.53  &0.40  & \textbf{0.57} & \textbf{0.43} \\
 \hline
150 &0.46  &0.39  & 0.51 &  0.41  \\
 \hline
\end{tabular}
\end{table}

\begin{table*}[h]
\caption{CCC and MSE based evaluation of valence \& arousal predictions provided by: 1) the CNN architecture when using three different pre-trained networks for
initialization (VGG-16, ResNet-50, VGG-Face) and 2) the VGG-Face-LSTM and AffWildNet architectures (2 RNN layers with 128 units each). A higher CCC and a lower MSE value indicate a better performance.}
\label{table1}
\parbox{.45\linewidth}{
\centering
\begin{tabular}{ |c||c|c|c| }
 \hline
 \multicolumn{1}{|c||}{} & \multicolumn{3}{c|}{CCC}  \\
  \hline
  & Valence & Arousal & Mean Value\\
 \hline
 VGG-16 &0.40 &0.30 & 0.35  \\
 \hline
 ResNet-50 &0.43  &0.30 & 0.37 \\
 \hline
VGG-Face & \textbf{0.51}& \textbf{0.33} & \textbf{0.42}      \\
 \hline
VGG-Face-LSTM & 0.52  &0.38 & 0.45  \\
\hline
 \textbf{AffWildNet} & \textbf{0.57}&  \textbf{0.43} & \textbf{0.50}      \\
 \hline
\end{tabular}
}
\hfil
\parbox{.45\linewidth}{
\centering
\begin{tabular}{ |c||c|c|c| }
 \hline
 \multicolumn{1}{|c||}{} & \multicolumn{3}{c|}{MSE}  \\
  \hline
  & Valence & Arousal & Mean Value\\
 \hline
 VGG-16 &0.13 &0.11 & 0.12  \\
 \hline
 ResNet-50 &0.11  &0.11 & 0.11  \\
 \hline
VGG-Face & \textbf{0.10}& \textbf{0.08} & \textbf{0.09}      \\
 \hline
VGG-Face-LSTM & 0.10  &0.09 & 0.10  \\
\hline
 \textbf{AffWildNet} & \textbf{0.08}&  \textbf{0.06} & \textbf{0.07}      \\
 \hline
\end{tabular}
}
\end{table*}

Table \ref{table1} summarizes the CCC and MSE values obtained when applying all developed architectures described in subsections \ref{cnn_dev} and \ref{cnn-rnn_dev},
to the Aff-Wild test set. 
It shows the improvement in the CCC and MSE values obtained when using the AffWildNet compared to all other developed architectures. This improvement clearly indicates the ability of the AffWildNet to better capture the dynamics in Aff-Wild. 

\begin{figure}[h]
\centering
\begin{subfigure}[Valence]
{\centering
\includegraphics[width=8.4cm,height=5cm,trim={0.5cm 0 0 0}]{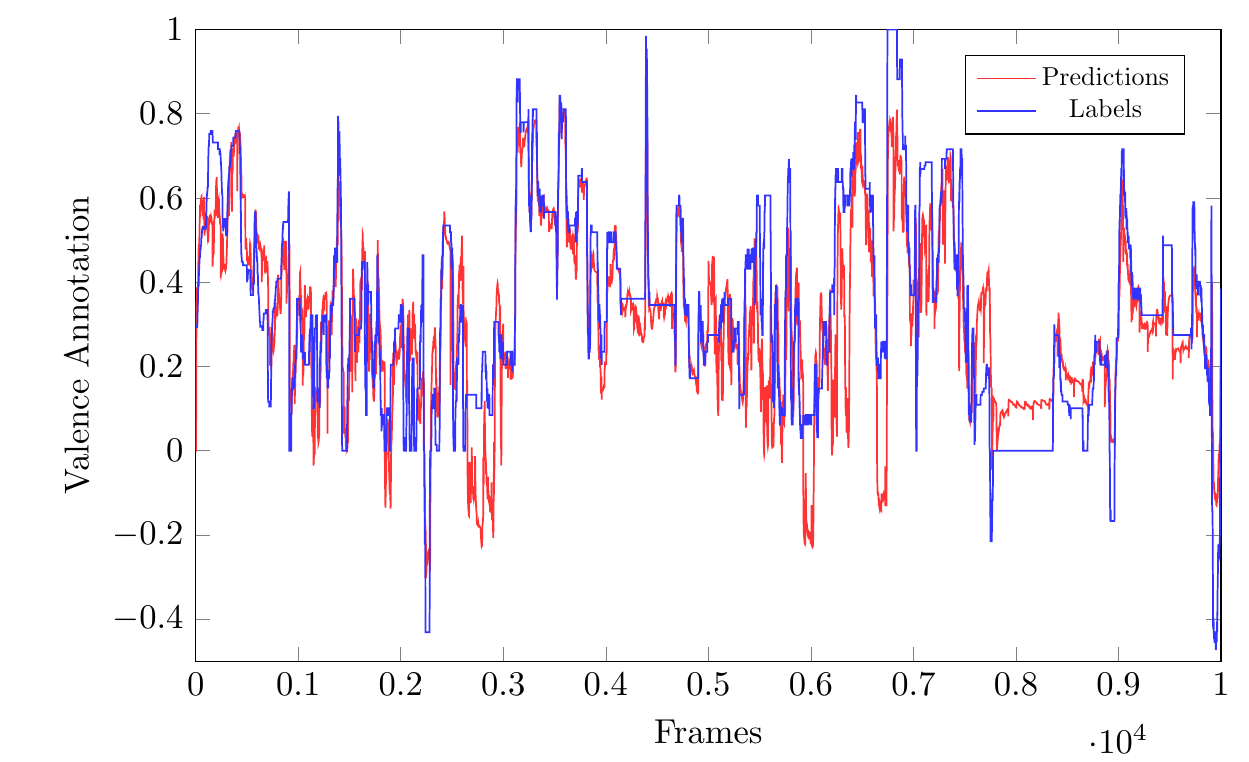}
\label{valence}
}
\end{subfigure}
\quad

\begin{subfigure}[Arousal]
{\centering
\includegraphics[width=8.4cm,height=5cm,trim={0.5cm 0 0 0}]{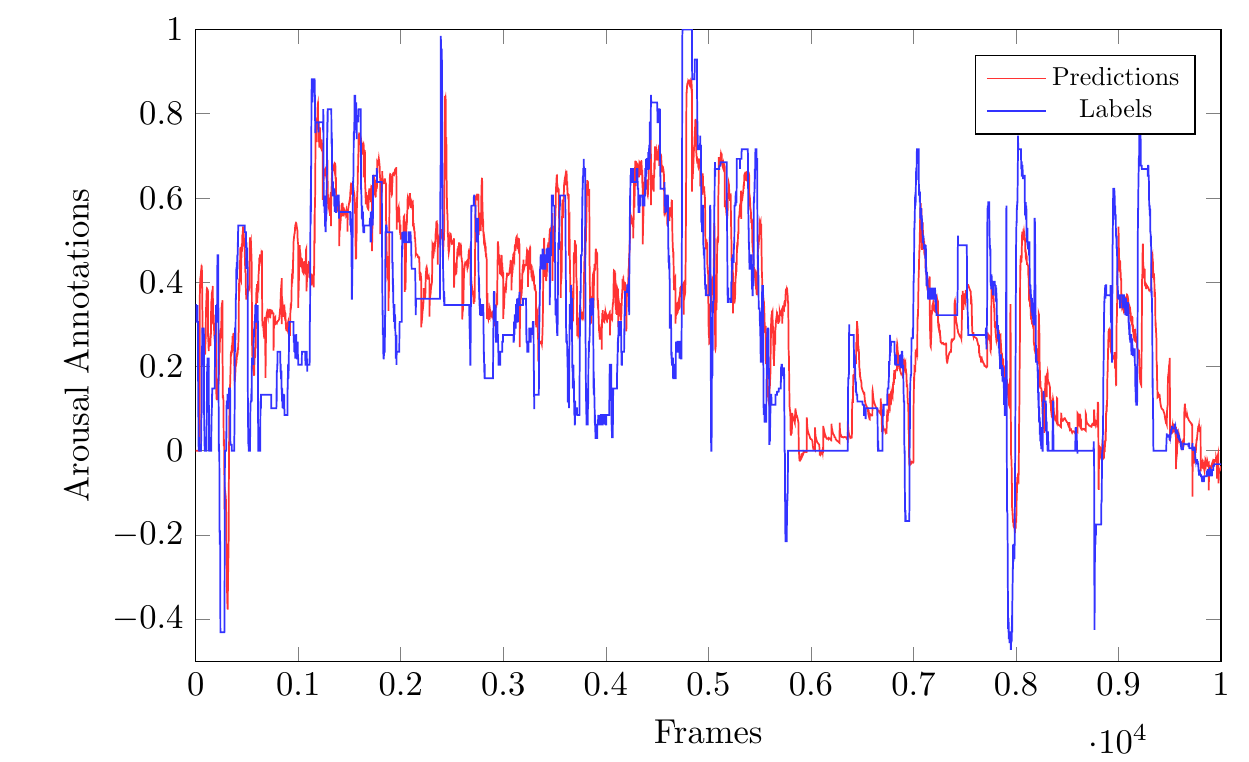}
\label{arousal}
}
\end{subfigure}
\caption{Predictions vs Labels for (a) valence and (b) arousal over a video segment of the Aff-Wild.}
\end{figure}

In Figures \ref{valence} and \ref{arousal}, we qualitatively illustrate some of the obtained results by comparing a segment of the obtained valence/arousal predictions to the ground truth values, in $10000$ consecutive frames of test data.

\begin{figure}[h]
\centering
\begin{subfigure}[annotations]
{\centering
\includegraphics[height=0.7\linewidth]{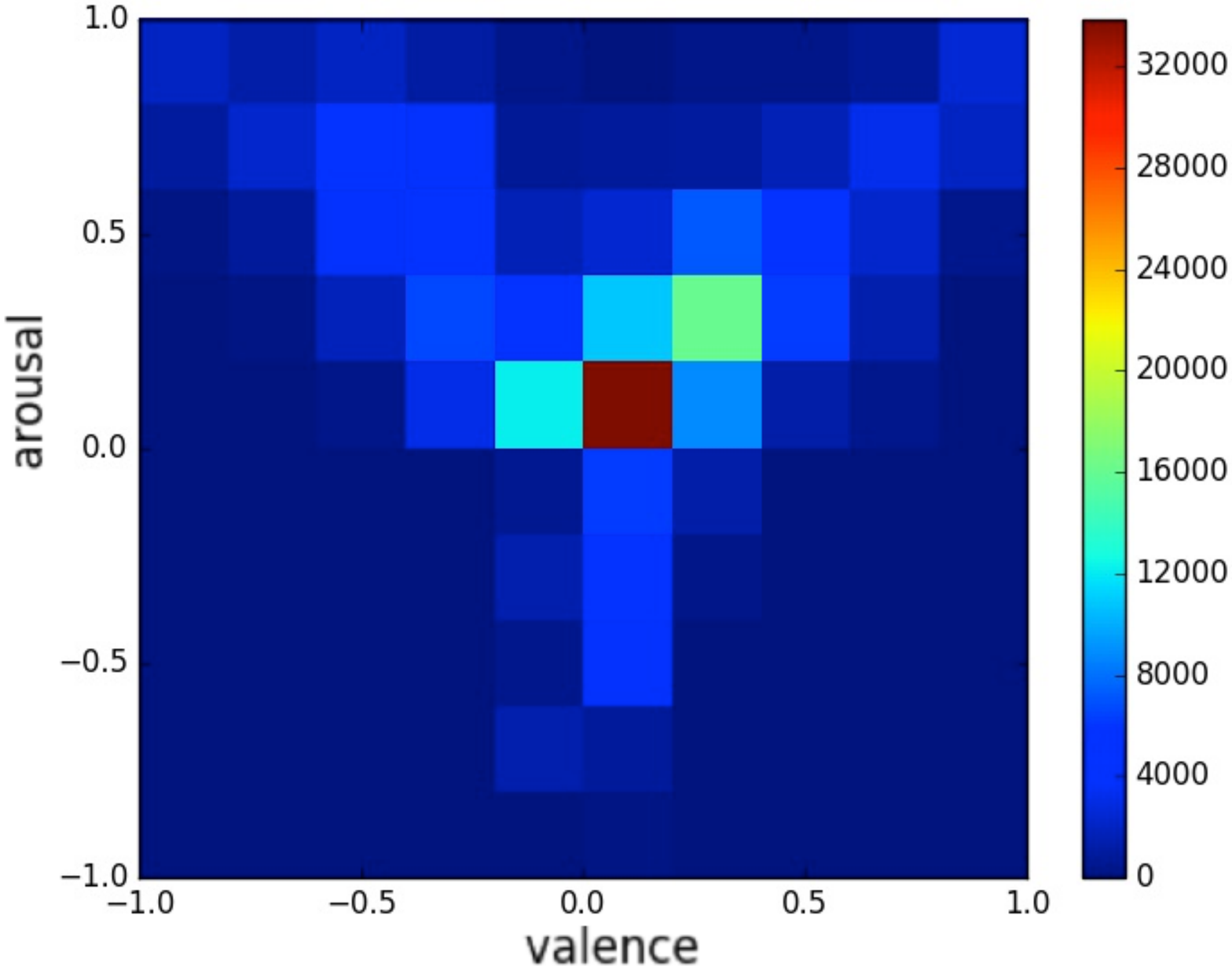}
\label{test_hist1}
}
\end{subfigure}
\quad
\begin{subfigure}[predictions]
{\centering
\includegraphics[height=0.7\linewidth]{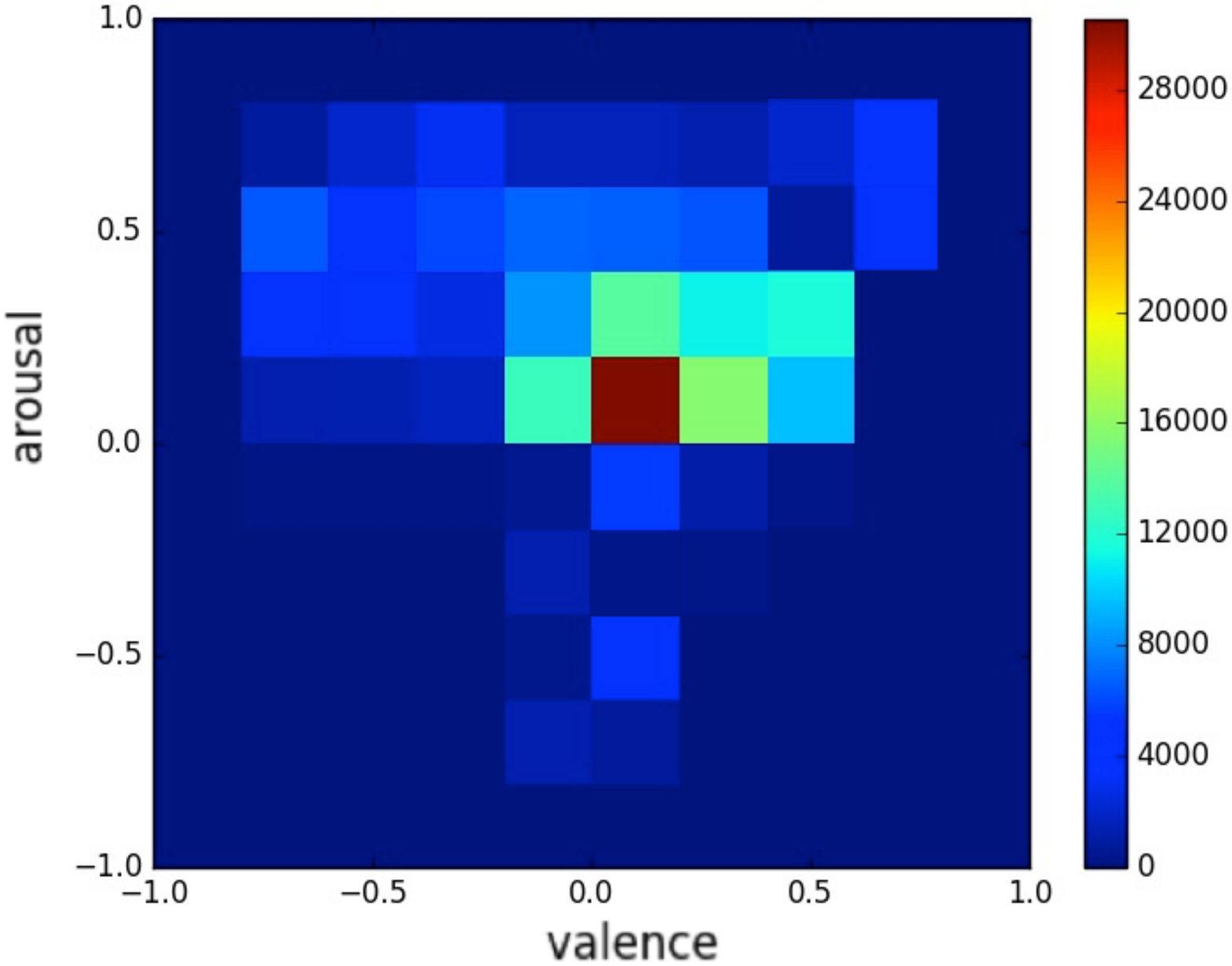}
\label{test_hist2}
}
\end{subfigure}

\caption{Histogram in the 2-D valence \& arousal space of: (a) annotations and (b) predictions of AffWildNet, on the test set of the Aff-Wild Challenge.}
\end{figure}

Moreover, in Figures \ref{test_hist1} and \ref{test_hist2}, we illustrate, in the 2-D valence \& arousal space, the histograms of the ground truth labels of the test set and the corresponding predictions of our AffWildNet.

The results shown in Table \ref{table1} and the above Figures verify the excellent performance of the AffWildNet. They also show that it greatly outperformed all methods submitted in the Aff-Wild Challenge.
\subsection{Discussing AffWildNet's Performance}

The reasons why the AffWildNet outperformed the other methods are related to both the network design and the network training.

At first, the AffWildNet is a CNN-RNN network. The CNN part is based on the VGG-Face (or ResNet-50) network's convolutional and pooling layers. The VGG-Face network has been pre-trained with a large dataset for face recognition (many human faces have been, therefore, used in its construction). 

In our implementation, this CNN part is followed by a single FC layer. The inputs of this layer are: a) the outputs of the last pooling layer of the CNN part; b) the facial landmarks, which are directly passed as inputs to this FC layer. As a consequence, this layer has the role to map its two types of inputs to the same feature space, before forwarding them to the RNN part. The facial landmarks, which are provided as additional input to the network, in this way, contribute to boosting the performance of our model.
The output of the fully connected layer is then passed to the RNN part.

The RNN is used in order to model the contextual information in the data, taking into account temporal variations. The RNN is composed of 2-layers, with GRU units in each layer; the first layer processes the FC layer outputs, the second layer is followed by the output layer that gives the final estimates for valence and arousal.

Part of AffWildNet's design was the fixing of its optimal hyper-parameters (number of FC and RNN layers, number of hidden units in these layers, batch size, sequence length, dropout, learning rate). Finally, the specification of the loss function used for network training was another important issue. Our loss function was based on the CCC, as this was the main evaluation criterion of the Aff-Wild Challenge; this was not the case in the competing methods that used the usual MSE criterion in their training phases.

As far as network training is concerned, the AffWildNet has been trained as an end-to-end architecture, by jointly training its CNN and RNN parts, rather than separately training the two parts.

We would also like to mention that the data augmentation that was conducted so as to achieve a more balanced dataset, also contributed in achieving the AffWildNet a state-of-the-art performance.

\section{Feature Learning from Aff-Wild}

When it comes to dimensional emotion recognition, there exists great variability between different databases, especially those containing emotions in-the-wild. In particular, the annotators and the range of the annotations are different and the labels can be either discrete or continuous. To tackle the problems caused by this variability, we take advantage of the fact that the Aff-Wild is a powerful database that can be exploited for learning features, which may then be used as priors for dimensional emotion recognition. In the following, we show that it can be used as prior for the RECOLA and AFEW-VA databases that are annotated for valence and arousal, just like Aff-Wild. In addition to this, we use it as a prior for categorical emotion recognition, on the EmotiW dataset, which is annotated in terms of the seven basic emotions. Experiments have been conducted on these databases yielding state-of-the-art results and thus verifying the strength of Aff-Wild for affect recognition.

\subsection{Prior for Valence and Arousal Prediction}

\subsubsection{Experimental Results for the Aff-Wild and RECOLA database}

In this subsection, we demonstrate the superiority  of our database when it is used for pre-training a DNN. In particular, we fine-tune the AffWildNet on the RECOLA and for comparison purposes we also train on RECOLA an architecture comprised of a ResNet-50 and a 2-layer GRU stacked on top (let us call it ResNet-GRU network). Table \ref{recola} shows the results only for the CCC score as our minimization loss was depending on this metric.
It is clear that the performance on both arousal and valence of the fine-tuned model on the Aff-Wild database is much higher than the performance of the ResNet-GRU model.

\begin{table}[!h]
\caption{CCC based evaluation of valence \& arousal predictions provided by the fine-tuned AffWildNet and the ResNet-GRU on the RECOLA test set. A higher CCC  value indicates a better performance.}
\label{recola}
\centering
\begin{tabular}{ |c||c|c| }
 \hline
 \multicolumn{1}{|c||}{} & \multicolumn{2}{c|}{CCC} \\
 \hline
     & Valence & Arousal  \\
 \hline
 \textbf{Fine-tuned AffWildNet} & \textbf{0.526} & \textbf{0.273}  \\
 \hline
 ResNet-GRU & 0.462 & 0.209  \\
 \hline
\end{tabular}
\end{table}

\begin{figure}[h]
\centering
\begin{subfigure}[annotations]
{\centering
\adjincludegraphics[height=0.7\linewidth]{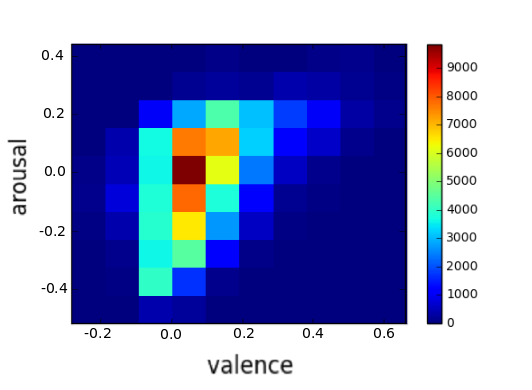}
\label{gth}
}
\end{subfigure}
\quad

\begin{subfigure}[predictions]
{\centering
\adjincludegraphics[height=0.7\linewidth]{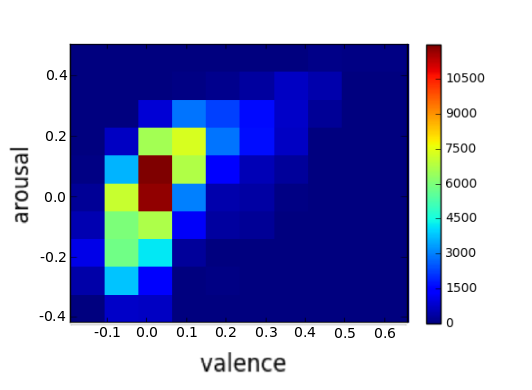} 
\label{preds}
}
\end{subfigure}

\caption{Histogram in the 2-D valence \& arousal space of (a) annotations and (b) predictions for the test set of the RECOLA database.}
\end{figure}

To further demonstrate the benefits of our model when predicting valence and arousal, we demonstrate a histogram in the 2-D valence \& arousal space of the annotations (Figure \ref{gth}) and predictions of the fine-tuned AffWildNet (Figure \ref{preds}) for the whole test set of RECOLA.

\begin{figure}[h]
\centering
\begin{subfigure}[Valence]
{\centering
\includegraphics[width=8.4cm,height=5cm,trim={0.8cm 0 0 0}]{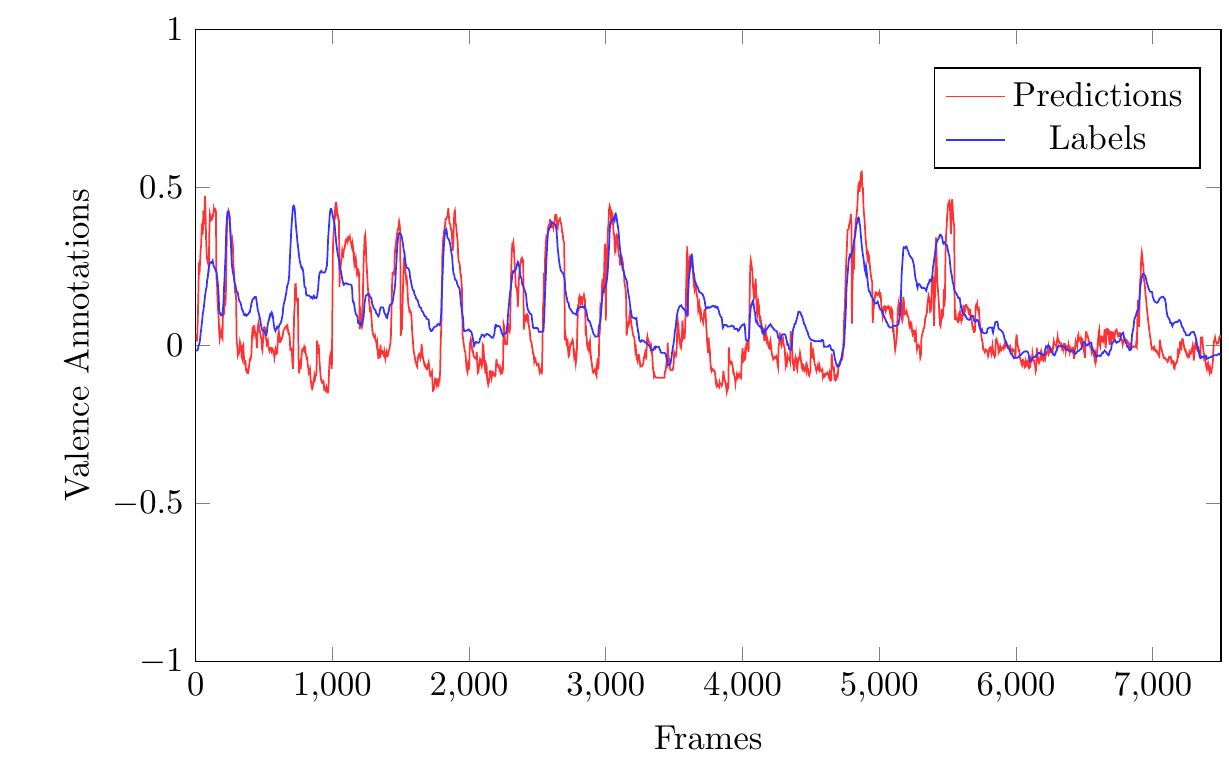}
\label{rvalence}
}
\end{subfigure}
\quad
\begin{subfigure}[Arousal]
{\centering
\includegraphics[width=8.4cm,height=5cm,trim={0.8cm 0 0 0}]{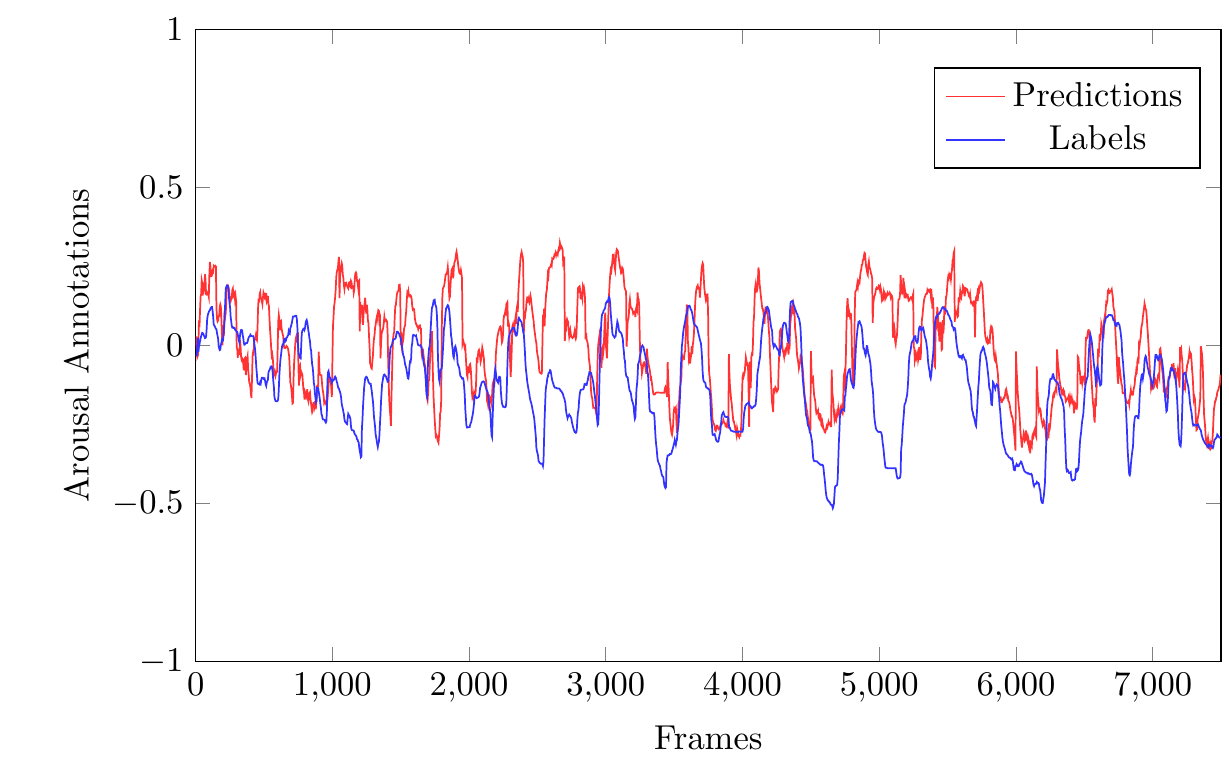}
\label{rarousal}
}
\end{subfigure}

\caption{Fine-tuned AffWildNet's Predictions vs Labels for (a) valence and (b) arousal for a single test video of the RECOLA database.}
\end{figure}

Finally, we also illustrate in Figures \ref{rvalence} and \ref{rarousal} the network prediction and ground truth for one test video of RECOLA, for the valence and arousal dimensions, respectively.

\subsubsection{Experimental Results for the AFEW-VA database}

\begin{figure}[h]
\begin{minipage}{.5\textwidth}
  \centering
  \includegraphics[height=0.7\linewidth]{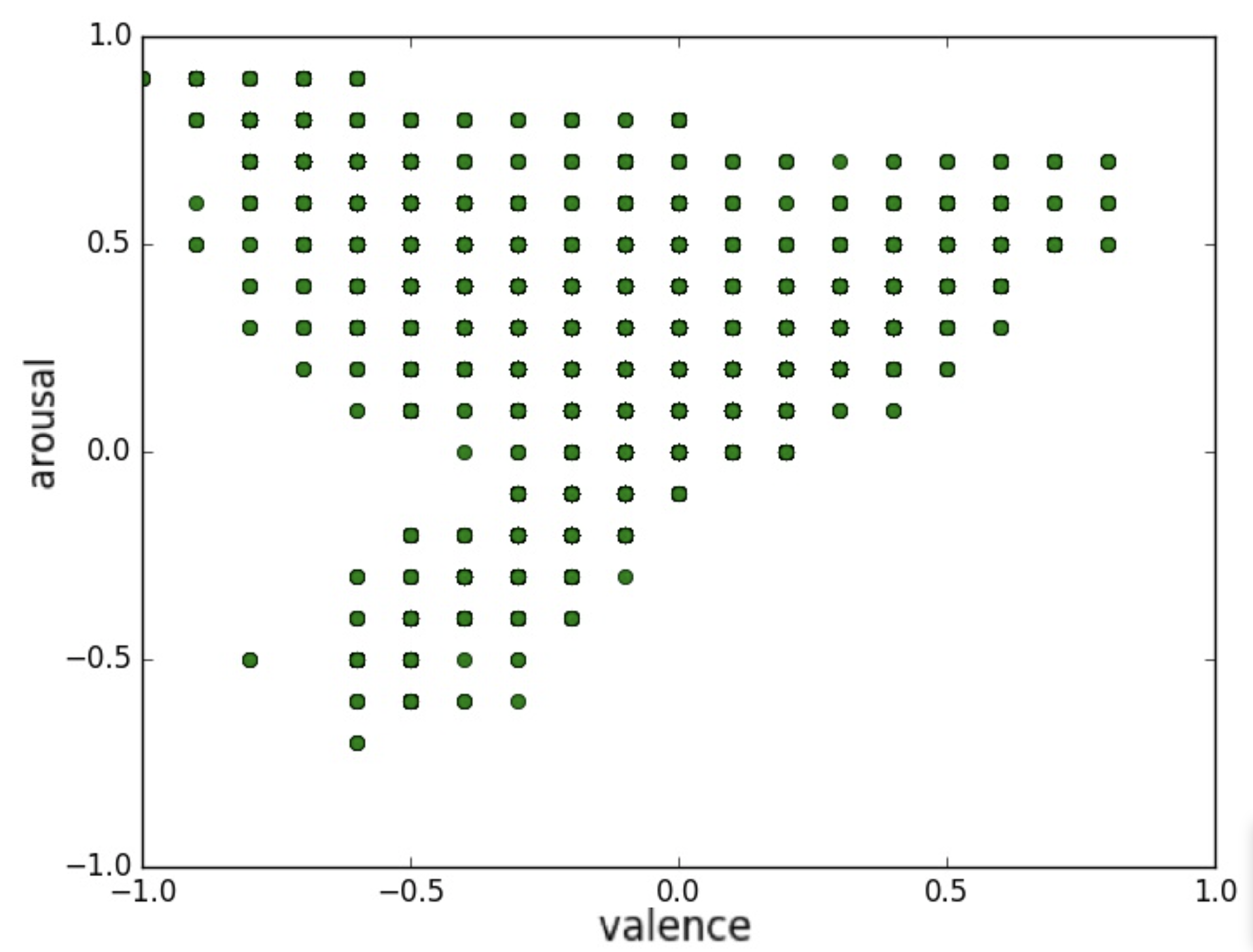}
\end{minipage}
\caption{Discrete values of annotations of the AFEW-VA database.}
\label{test_afewva_labels1}
\begin{minipage}{.4\textwidth}
  \centering
  \includegraphics[height=0.95\linewidth]{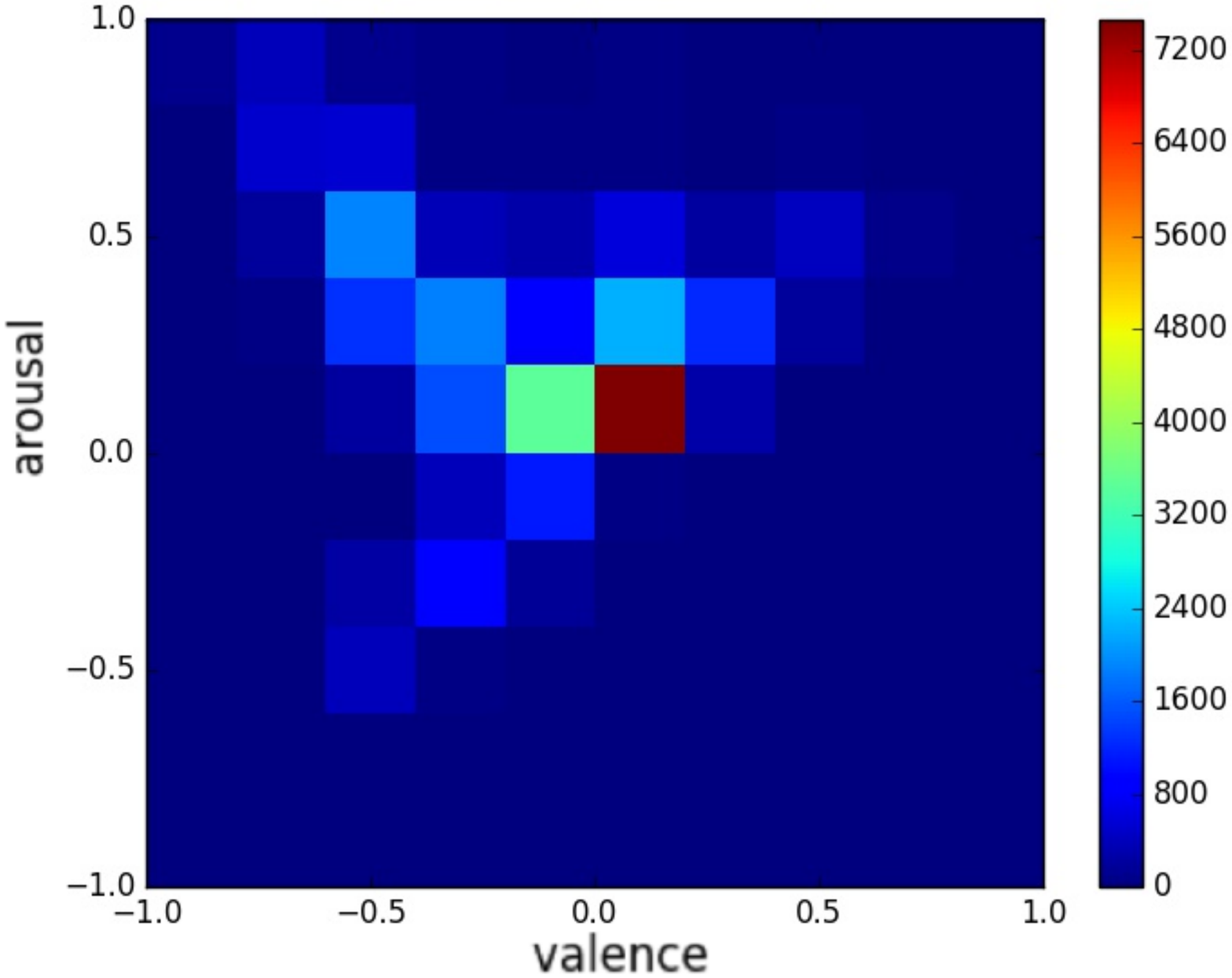}
\end{minipage}
\caption{Histogram in the 2-D valence \& arousal space of annotations of the AFEW-VA database.}
\label{test_afewva_labels2}
\end{figure}

In this subsection, we focus on recognition of emotions in the AFEW-VA database, which annotation's is somewhat different from the annotation of the Aff-Wild database. In particular, the labels of the AFEW-VA database are in the range [$-10 $, $+10 $], while the labels of the Aff-Wild database are in the range [$-1 $, $+1 $]. To tackle this problem, we scaled the range of the AFEW-VA labels to [$-1 $, $+1 $]. Moreover, differences were observed, due to the fact that the labels of the AFEW-VA are discrete, while the labels of the Aff-Wild are continuous. Figure \ref{test_afewva_labels1} shows the discrete valence and arousal values of the annotations in AFEW-VA database, whereas Figure \ref{test_afewva_labels2} shows the corresponding histogram in the 2-D valence \& arousal space.

We then performed fine-tuning of the AffWildNet to the AFEW-VA database and tested the performance of the generated network. Similarly to \cite{kossaifi2017afew}, we used a 5-fold person-independent cross-validation strategy. Table \ref{affewvacomp} shows a comparison of the performance of the fine-tuned AffWildNet with the best results reported in \cite{kossaifi2017afew}. Those results are in terms of the Pearson CC. It can be easily seen that the fine-tuned AffWildNet greatly outperformed the best method reported in \cite{kossaifi2017afew}.

\begin{table}[h]
\caption{Pearson Correlation Coefficient (Pearson CC) based evaluation of valence \& arousal predictions provided by the best architecture in \cite{kossaifi2017afew} vs our AffWildNet fine-tuned on the AFEW-VA. A higher Pearson CC value indicates a better performance.}
\label{affewvacomp}
\centering
\begin{tabular}{ |c||c|c|  }
 \hline
 \multicolumn{1}{|c||}{Group} & \multicolumn{2}{c|}{Pearson CC}   \\
 \hline
     & Valence & Arousal  \\
\hline
best of \cite{kossaifi2017afew} & 0.407 & 0.45 \\
\hline
\textbf{Fine-tuned AffWildNet} & \textbf{0.514} & \textbf{0.575}    \\
 \hline
\end{tabular}
\end{table}

\noindent

For comparison purposes, we also trained a CNN network on the AFEW-VA database. This network's architecture was based on the convolution and pooling layers of VGG-Face followed by 2 fully connected layers with 4096 and 2048 hidden units, respectively. As shown in Table \ref{affewva}, the performance of the fine-tuned AffWildNet, in terms of CCC, greatly outperformed this network as well.

\begin{table*}[h]
\caption{Accuracies on the EmotiW validation set obtained by different CNN and CNN-RNN architectures vs the fine-tuned AffWildNet. A higher accuracy value indicates better performance.}
\label{afew}
\centering
\begin{tabular}{|c||c|c|c|c|c|c|c|c|}
\hline
\multicolumn{1}{|c||}{Architectures} & \multicolumn{8}{c|}{Accuracy}  \\
\hline
& Neutral & Anger & Disgust & Fear & Happy & Sad & Surprise & Total \\
\hline
VGG-16 & 0.327 & 0.424 & \textbf{0.102} & \textbf{0.093}  & 0.476 & 0.138  & \textbf{0.133}  & 0.263 \\
\hline
VGG-16 + RNN & 0.431 & 0.559 & 0.026 & 0.07  & 0.444 & 0.259  & 0.044  & 0.293 \\
\hline
ResNet & 0.31 & 0.153 & 0.077 & 0.023  & 0.534 & 0.207  & 0.067  & 0.211 \\
\hline
ResNet + RNN & 0.431 & 0.237 & 0.077 & 0.07  & 0.587 & 0.155  & 0.089  & 0.261 \\
\hline
VGG-Face + RNN & 0.552 & 0.593 & 0.026 & 0.047  & \textbf{0.794} & 0.259  & 0.111  & 0.384 \\
\hline
\textbf{fine-tuned AffWildNet} & \textbf{0.569} & \textbf{0.627} & 0.051 & 0.023  & 0.746 & \textbf{0.709}  & 0.111  & \textbf{0.454} \\
\hline
\end{tabular}
\end{table*}

\begin{table}[h]
\caption{CCC based evaluation of valence \& arousal predictions provided by the CNN architecture based on VGG-Face and the fine-tuned AffWildNet on the AFEW-VA training set. A higher CCC  value indicate a better performance.}
\label{affewva}
\centering
\begin{tabular}{ |c||c|c|  }
 \hline
 \multicolumn{1}{|c||}{} & \multicolumn{2}{c|}{CCC AFEW-VA} \\ 
 \hline
     & Valence & Arousal  \\
 \hline
only CNN & 0.44& 0.474       \\
\hline
\textbf{Fine-tuned AffWildNet}  & \textbf{0.515} & \textbf{0.556}       \\
 \hline
\end{tabular}
\end{table}

All these verify that our network can be used as a pre-trained one to yield excellent results across different dimensional databases.

\subsection{Prior for Categorical Emotion Recognition}

\subsubsection{Experimental Results for the EmotiW dataset}

To further show the strength of the AffWildNet, we used the AffWildNet - which is trained for dimensional emotion recognition task - in a very different problem, that of  
categorical in-the-wild emotion recognition, focusing on the EmotiW 2017 Grand Challenge. To tackle categorical emotion recognition, we modified the AffWildNet's output layer to include 7 neurons (one for each basic emotion category) and performed fine-tuning on the AFEW 5.0 dataset. 

In the presented experiments, we compare the fine-tuned AffWildNet's performance with that of other state-of-the-art CNN and CNN-RNN networks; the CNN part of which is based on the ResNet 50, VGG-16 and VGG-Face architectures, trained on the same AFEW 5.0 dataset. The accuracies of all networks on the validation set of the EmotiW 2017 Grand Challenge are shown in Table \ref{afew}. A higher accuracy value indicates better performance for the model. We can easily see that the AffWildNet outperforms all those other networks in terms of total accuracy.

\noindent
We should  note that:

\begin{itemize}
\item[(i)]  the AffWildNet was trained to classify only video frames (and not audio) and then video classification based on frame aggregation was performed
\item[(ii)] the cropped faces provided by the challenge were only used (and not our own detection and/or normalization procedure)
\item[(iii)] no data-augmentation, post-processing of the results or ensemble methodology have been conducted. 
\end{itemize}

\noindent
It should also be mentioned that the fine-tuned AffWildNet's performance, in terms of total accuracy, is:
\begin{itemize}
\item[(i)]  much higher than the baseline total accuracy of 0.3881 reported in \cite{dhall2017individual}
\item[(ii)] better than all vanilla architectures' performances that were reported by the three winning methods in the audio-video emotion recognition EmotiW 2017 Grand Challenge \cite{hu2017learning} \cite{knyazev2017convolutional} \cite{vielzeuf2017temporal}
\item[(iii)] comparable and better in some cases than the rest of the results obtained by the three winning methods \cite{hu2017learning} \cite{knyazev2017convolutional} \cite{vielzeuf2017temporal}
\end{itemize}

\begin{table*}[h]
\caption{Overall accuracies of the best architectures of the three winning methods of the EmotiW 2017 Grand Challenge reported on the validation set vs our fine-tuned AffWildNet. A higher accuracy value indicates better performance.}
\label{emotiw}
\centering
\begin{tabular}{|c||c|c|c|c|}
\hline
\multicolumn{1}{|c||}{Group} & \multicolumn{1}{c|}{Architecture} & \multicolumn{3}{c|}{Total Accuracy} \\
\hline
& & Original & \begin{tabular}{@{}c@{}}After\\ Fine-Tuning \\on FER2013 \end{tabular}  & \begin{tabular}{@{}c@{}}Data\\ augmentation\end{tabular} \\
\hline
\cite{hu2017learning} &  \begin{tabular}{@{}c@{}} DenseNet-121 \\ HoloNet \\ ResNet-50 \end{tabular}   &   \begin{tabular}{@{}c@{}}  0.414 \\ 0.41 \\ 0.418 \end{tabular}  & - & -\\
\hline
\cite{knyazev2017convolutional} &  \begin{tabular}{@{}c@{}} VGG-Face \\ FR-Net-A \\ FR-Net-B \\ FR-Net-C \\ LSTM + FR-NET-B \end{tabular}    &  \begin{tabular}{@{}c@{}} 0.379 \\ 0.337 \\ 0.334 \\ 0.376 \\ - \end{tabular} & \begin{tabular}{@{}c@{}}  0.483 \\ 0.446 \\ \textbf{0.488} \\  0.452 \\ 0.465 \end{tabular} &\begin{tabular}{@{}c@{}} - \\ - \\ - \\ - \\ \textbf{0.504} \end{tabular} \\
\hline
\cite{vielzeuf2017temporal}& \begin{tabular}{@{}c@{}} Weighted C3D (no overlap) \\ LSTM C3D (no overlap) \\ VGG-Face \\ VGG-LSTM 1 layer \end{tabular} & - & - & \begin{tabular}{@{}c@{}} 0.421 \\ 0.432 \\ 0.414 \\ 0.486 \end{tabular}\\
\hline
Our & Fine-tuned AffWildNet   & \textbf{0.454} & - & - \\
\hline
\end{tabular}
\end{table*}

\noindent
The above are shown in Table \ref{emotiw}. Those results verify that the AffWildNet can be appropriately fine-tuned and successfully used for dimensional, as well as for categorical emotion recognition.

\section{Conclusions and Future Work}

Deep learning and deep neural networks have been successfully used in the past years for facial expression and emotion recognition based on still image and video frame analysis. Recent research focuses on in-the-wild facial analysis and refers either to categorical emotion recognition, targeting recognition of the seven basic emotion categories, or to dimensional emotion recognition, analyzing the valence-arousal (V-A) representation space.

In this paper, we introduce Aff-Wild, a new, large in-the-wild database that consists of 298 videos of 200 subjects, with a total length of more than 30 hours. We also present the Aff-Wild Challenge that was organized on Aff-Wild. We report the results of the challenge, and the pitfalls and challenges in  terms of predicting valence and arousal in-the-wild. Furthermore, we design a deep convolutional and recurrent neural architecture and perform extensive experimentation with the Aff-Wild database. We show that the generated AffWildNet provides the best performance for valence and arousal estimation on the Aff-Wild dataset, both in terms of the Concordance Correlation Coefficient and  the Mean Squared Error criteria, when compared with other deep learning networks trained on the same database.

Subsequently, we then demonstrate that the AffWildNet and Aff-Wild database constitute tools that can be used for facial expression and emotion recognition on other datasets. Using appropriate fine-tuning and retraining methodologies, we show that best results can be obtained by applying the AffWildNet to other dimensional databases, including the RECOLA and the AFEW-VA ones and by comparing the obtained performances with other state-of-the-art pre-trained and fine-tuned networks.

Furthermore, we observe that fine-tuning on the AffWildNet can produce state-of-the-art performance, not only for dimensional, but also for categorical emotion recognition. We use this approach to tackle the facial expression and emotion recognition parts of the EmotiW 2017 Grand Challenge, referring to recognition of the seven basic emotion categories, finding that we produce comparable or better results to the winners of this contest.

It should be stressed that it is the first time, to the best of our knowledge, that the same deep architecture can be used for both types of dimensional and categorical emotion analysis. To achieve this, the AffWildNet has been effectively trained with the largest existing, in-the-wild, database for continuous valence-arousal recognition (regression analysis problem) and then used for tackling the discrete seven basic emotion recognition (classification) problem.

The proposed procedure for fine-tuning the AffWildNet can be applied to further extend its use in the analysis of other new visual emotion recognition datasets. This includes our current work on extending the Aff-Wild with new in-the-wild audiovisual information, as well as using it as a means for unifying different approaches
to facial expression and emotion recognition. These approaches contain   dimensional emotion representations, basic and compound emotion categories, facial action unit representations, as well as specific emotion categories met in different contexts, such as negative emotions, emotions in games, in social groups and other human machine (or robot) interactions.

\begin{acknowledgements}
The work of Stefanos Zafeiriou has been partially funded by the FiDiPro program of Tekes with project number 1849/31/2015. The work of Dimitris Kollias was funded by a Teaching Fellowship of Imperial College London. The support of the EPSRC Centre for Doctoral Training in High Performance Embedded and Distributed Systems (HiPEDS, Grant Reference EP/L016796/1) is gratefully acknowledged. We also
thank the NVIDIA Corporation for donating a Titan X GPU.  We would like also to acknowledge the contribution of the Youtube users that gave us the permission to use their videos (especially Zalzar and Eddie from The1stTake). We wish to thank Dr A Dhall for providing us with the data of the Emotiw 2017 Grand Challenge. Additionally, we would like to thank the reviewers for their  valuable comments that helped us to improve this paper.
\end{acknowledgements}

\bibliographystyle{spmpsci}      
\bibliography{sample}   

\appendix
\section{Appendix}

\subsection{Baseline: CNN-M}

The exact structure of the network is shown in Table \ref{baseline}. In total, it consists of 5 convolutional, batch normalization and pooling layers and 2 fully connected (FC) ones. For each convolutional layer the parameters are the filter and the stride, in the form of (filter height, filter width, input channels , output channels/feature maps) and (1, stride height, stride width , 1), respectively, and for the max pooling layer the parameters are the ksize and stride, in the form of (pooling height, pooling width, input channels, output channels) and (1, stride height, stride width , 1), respectively. We follow the TensorFlow's platform notation for the values of all those parameters. Note that the activation function in the convolutional and batch normalization layers is the ReLU one; this is also the case in the first FC layer. The activation function of the second FC layer, which is the output layer, is a linear one.

\begin{table}[h]
\centering
\captionsetup{font=small}
\caption{ Baseline architecture based on CNN-M, showing the values of the parameters of the convolutional and pooling layers and the number of hidden units in the fully connected layers. We follow the TensorFlow's platform notation for the values of all those parameters.}
\label{baseline}
\scalebox{0.73}{
\begin{tabular}{|c|c|c|c|c|c|}
\hline
Layer & filter & ksize & stride & padding & no of units \\
\hline
conv 1 & [7, 7, 3, 96] & & [1, 2, 2, 1] & 'VALID' & \\
batch norm& & & & &\\
max pooling & & [1, 3, 3, 1] & [1, 2, 2, 1] & 'VALID' & \\
\hline
conv 2 & [5, 5, 96, 256] & & [1, 2, 2, 1] & 'SAME' & \\
batch norm& & & & &\\
max pooling & & [1, 3, 3, 1] & [1, 2, 2, 1] & 'SAME' & \\
\hline
conv 3 & [3, 3, 256, 512] & & [1, 1, 1, 1] & 'SAME' & \\
batch norm& & & & &\\
\hline
conv 4 & [3, 3, 512, 512] & & [1, 1, 1, 1] & 'SAME' & \\
batch norm& & & & &\\
\hline
conv 5 & [3, 3, 512, 512] & & [1, 1, 1, 1] & 'SAME' & \\
batch norm& & & & &\\
max pooling & & [1, 2, 2, 1] & [1, 2, 2, 1] & 'SAME' & \\
\hline
fully connected 1 &&&&&4096\\
\hline
fully connected 2 &&&&&2\\
\hline
\end{tabular}}
\vspace{-6mm}
\end{table}

\subsection{ResNet-50}

\begin{figure*}
\centering
\adjincludegraphics[width=15cm]{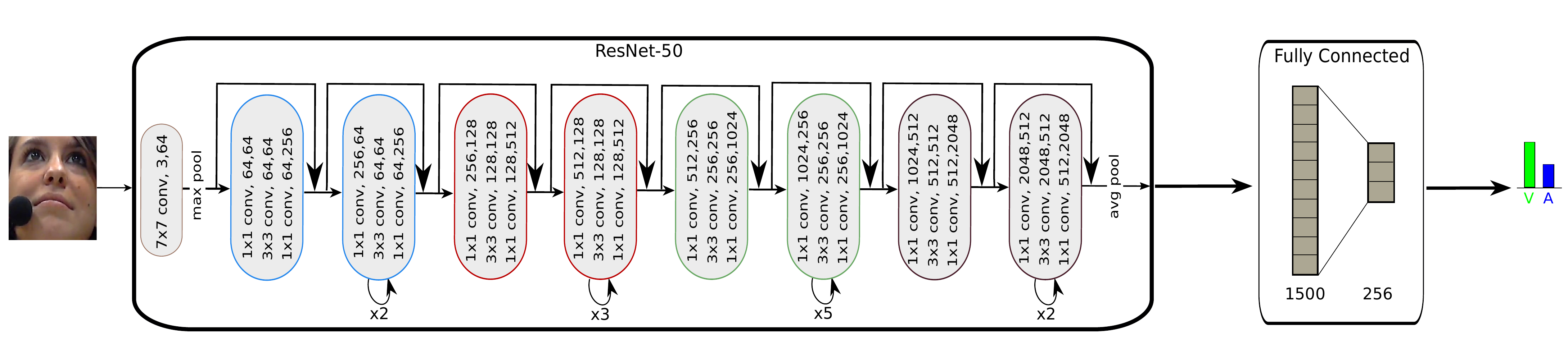} 
\caption{The CNN-only architecture for valence and arousal estimation, based on ResNet-50 structure and including two fully connected layers (V and A stand for valence and arousal respectively). Each convolutional layer is in the format: filter height $\times$ filter width, number of input feature maps, number of output feature maps.}
\label{ResNet_FC}
\end{figure*}

\begin{figure*}
\centering
\adjincludegraphics[height=5.5cm,width=14cm]{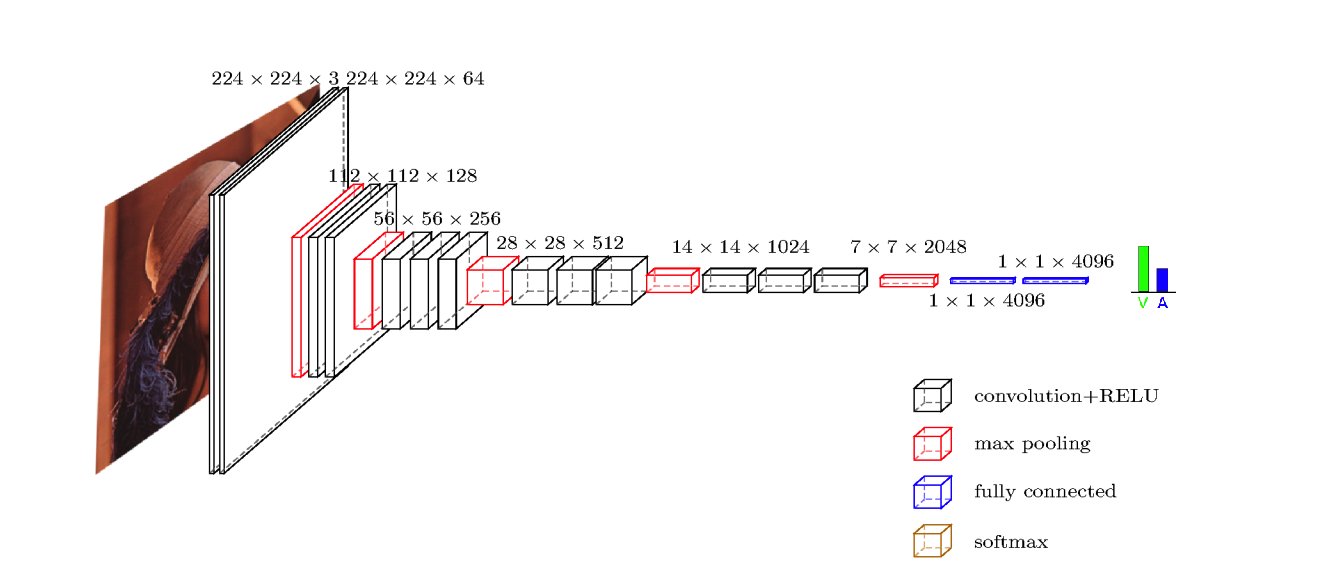} 
\caption{The CNN-only architecture for valence and arousal estimation, based on VGG-Face structure (V and A stand for valence and arousal respectively).}
\label{vgg16}
\end{figure*}

Residual learning is adopted in these models by stacking multiple blocks of the form:

\begin{equation}
\mathbf{o}_k = \mathcal{B}(\mathbf{x}_k,\mathbf{\{W}_k\}) + h(\mathbf{x}_k),
\end{equation}

\noindent
where $\mathbf{x}_k$, $\mathbf{W}_k$ and $\mathbf{o}_k$ indicate the input, the weights, and the output of layer $k$, respectively, $\mathcal{B}$ indicates the residual function that is learnt and $h$ is the identity mapping between the residual function and the input. The $h$ identity mapping is a projection of $\mathbf{x}_k$ to match the dimensions of $\mathcal{B}(\mathbf{x}_k,\mathbf{\{W}_k\})$ (done by $1 \times 1$ convolutions), as in \cite{he2016deep}.

The first layer of the ResNet-50 model is comprised of a $7 \times 7$ convolutional layer with 64 feature maps, followed by a max pooling layer of size $3 \times 3$. Next, there are 4-bottleneck blocks, where a shortcut connection is added after each block. Each of these blocks is comprised of  3 convolutional layers of sizes $1 \times 1$, $3 \times 3$, and $1 \times 1$ with different number of feature maps. 

The architecture of the network is depicted in Figure \ref{ResNet_FC}.  Each convolutional layer is in the format: filter height $\times$ filter width, number of input feature maps, number of output feature maps.

\subsection{VGG-Face/VGG-16}

Table \ref{VGG16} shows the configuration of the CNN architecture based on VGG-Face or VGG-16. In total, it is composed of thirteen convolutional and pooling layers and three fully connected ones. For all those layers the form of the parameters is the same as described above in the baseline architecture.
We follow the TensorFlow's platform notation for the values of all those parameters. The output number of units is also shown in the Table. 

A linear activation function was used in the last FC layer, providing the final estimates. All units in the remaining FC layers were equipped with the ReLU. Dropout has been added after the first FC layer in order to avoid over-fitting. The architecture of the network is depicted in Figure \ref{vgg16}.

\begin{table}[h]
\centering
\captionsetup{font=small}
\caption{CNN architecture based on VGG-Face/VGG-16, showing the values of the parameters of the convolutional and pooling layers and the number of hidden units in the fully connected layers. We follow the TensorFlow's platform notation for the values of all those parameters.}
\label{VGG16}
\scalebox{0.75}{
\begin{tabular}{|c|c|c|c|c|c|}
\hline
Layer & filter & ksize & stride & padding & no of units \\
\hline
conv 1 & [3, 3, 3, 64] & & [1, 1, 1, 1] & 'SAME' & \\
\hline
conv 2 & [3, 3, 64, 64] & & [1, 1, 1, 1] & 'SAME' & \\
max pooling & & [1, 2, 2, 1] & [1, 2, 2, 1] & 'SAME' & \\
\hline
conv 3 & [3, 3, 64, 128] & & [1, 1, 1, 1] & 'SAME' & \\
\hline
conv 4 & [3, 3, 128, 128] & & [1, 1, 1, 1] & 'SAME' & \\
max pooling & & [1, 2, 2, 1] & [1, 2, 2, 1] & 'SAME' & \\
\hline
conv 5 & [3, 3, 128, 256] & & [1, 1, 1, 1] & 'SAME' & \\
\hline
conv 6 & [3, 3, 256, 256] & & [1, 1, 1, 1] & 'SAME' & \\
\hline
conv 7 & [3, 3, 256, 256] & & [1, 1, 1, 1] & 'SAME' & \\
max pooling & & [1, 2, 2, 1] & [1, 2, 2, 1] & 'SAME' & \\
\hline
conv 8 & [3, 3, 256, 512] & & [1, 1, 1, 1] & 'SAME' & \\
\hline
conv 9 & [3, 3, 512, 512] & & [1, 1, 1, 1] & 'SAME' & \\
\hline
conv 10 & [3, 3, 512, 512] & & [1, 1, 1, 1] & 'SAME' & \\
max pooling & & [1, 2, 2, 1] & [1, 2, 2, 1] & 'SAME' & \\
\hline
conv 11 & [3, 3, 512, 512] & & [1, 1, 1, 1] & 'SAME' & \\
\hline
conv 12 & [3, 3, 512, 512] & & [1, 1, 1, 1] & 'SAME' & \\
\hline
conv 13 & [3, 3, 512, 512] & & [1, 1, 1, 1] & 'SAME' & \\
max pooling & & [1, 2, 2, 1] & [1, 2, 2, 1] & 'SAME' & \\
\hline
fully connected 1 &&&&&4096\\
dropout &&&&& \\
\hline
fully connected 2 &&&&&4096\\
\hline
fully connected 3 &&&&&2\\
\hline
\end{tabular}}
\end{table}

\end{document}